\documentclass[10pt,twocolumn,letterpaper]{article}

\usepackage[pagenumbers]{cvpr} 

\usepackage{graphicx}
\usepackage{comment}
\usepackage{amsmath,amssymb} 
\usepackage{color}

\usepackage{bm} 

\DeclareMathOperator{\sgn}{sgn}

\usepackage{footmisc}
\renewcommand{\Re}[1]{{\operatorname{Re}\{#1\}}}

\DeclareMathOperator{\Tr}{Tr}
\DeclareMathOperator*{\argmax}{argmax} 
\newcommand{\br}{\boldsymbol{\mathrm{r}}}

\newcommand{\bn}{\boldsymbol{\hat{n}}}       
\newcommand{\bS}{\boldsymbol{\mathrm{S}}}
\newcommand{\bST}{\boldsymbol{\mathrm{ST}}}

\newcommand{\bom}{\boldsymbol{\omega}} 

\newcommand{\T}{\mathsf{T}} 

\def\Limsize{\columnwidth}

\newlength{\fgds}
\newlength{\fgdm}
\newlength{\fgdl}
\mathchardef\mhyphen="2D
\newcommand\ND{\mathop{{N{\mhyphen}D}}}
\newcommand\OND{\mathop{{1{\mhyphen}D}}}
\newcommand\TWD{\mathop{{2{\mhyphen}D}}}
\newcommand\THD{\mathop{{3{\mhyphen}D}}}

\usepackage{paralist}

\usepackage[font=small,labelfont=bf]{caption}
\definecolor{cvprblue}{rgb}{0.21,0.49,0.74}
\usepackage[pagebackref,breaklinks,colorlinks,allcolors=cvprblue]{hyperref}

\newcommand{\titletext}{
Total Least Square Optimal Analytic Signal by Structure Tensor for N-D
images}

\title{\strut
  {\titletext}\thanks{Due to presence
  of color-coded information with fine details, this paper is most appropriately viewed on a screen with  software having
  access to image zoom,
   or on print in color at high resolution.}
}

\author{Josef Bigun \quad Fernando Alonso-Fernandez\\
Halmstad University\\
SE-30118 Halmstad Sweden\\
{\tt\small josef.bigun@hh.se} \quad  {\tt\small Fernando.Alonso-Fernandez@hh.se}
}
    
\begin{document}

\maketitle
\thispagestyle{empty}

\begin{center}
\textit{This is a preprint and has not been peer-reviewed, including its previous version.}
\end{center}

\begin{abstract}
We produce the analytic signal by using the Structure
Tensor, which provides Total Least Squares optimal vectors
for estimating orientation and scale locally.
Together, these vectors represent N-D frequency components that determine
adaptive, complex probing filters.
The N-D analytic signal is obtained through scalar
products of adaptive filters with image neighborhoods. It comprises orientation,
scale, phase, and amplitude information of the neighborhood.
The ST analytic signal $ f_A $ is continuous and isotropic, and its
extension to N-D is straightforward. The phase gradient can be
represented as a vector (instantaneous frequency) or as a tensor. Both
are continuous and isotropic, while the tensor additionally preserves
continuity of orientation and retains the same information as the
vector representation. The tensor representation can also be used to
detect singularities. Detection with known phase portraits has been
demonstrated in 2-D with relevance to fringe pattern processing in
wave physics, including optics and fingerprint measurements.
To construct adaptive filters we have
used Gabor filter family members as probing functions, but other
function families can also be used to sample the spectrum, e.g.,
quadrature filters. A comparison to three baseline alternatives—in representation (Monogenic signal), enhancement (Monogenic signal combined with a spline-wavelet pyramid), and singularity detection (mindtct, a fingerprint minutia detector widely used in numerous studies)—is also reported using images with precisely known ground truths for location, orientation, singularity type (where applicable), and wave period.
\end{abstract} 

\setlength{\textfloatsep}{5pt plus 1.0pt minus 1.0pt}
\setlength{\floatsep}{3.0pt plus 1.0pt minus 1.0pt}
\setlength{\intextsep}{ 3.0pt plus 1.0pt minus 1.0pt}

\section{%
Introduction%
}
Starting with  phase~\cite{Sanger88,Jepson89,Wilson89,wiskott1997face,Hemmendorff99}, approximations of the analytic signal have long been used in image processing, typically through Gabor filters.
A known limitation of this approach is that it yields multiple amplitudes and phases per image point, which must be reconciled.
The term analytic signal originates from the study of real-valued
functio ns   $f(x)$,  $ x $ defined on $ \mathbb{R} $, where the negative frequency components and the DC component are suppressed in the Fourier domain~\cite{gabor}. This operation yields a unique complex-valued signal whose real part is $ f(x) $.  Its imaginary part is constructed such that the Fourier transform of the complex signal vanishes for all negative frequencies and at zero.

For real functions $f(\br)$ with $\br \in \ND$, however, this uniqueness no longer holds, since there are infinitely many ways to partition the $\ND$ frequency space into halves.
Our analytic signal is therefore constructed with the same underlying principle in mind: the spectrum $F(\bom)$ of $f$ is probed over only one half of the $\bom$-space by an adaptive, complex spatial filter that best explains the local image spectrum—without explicitly sampling it at every image location. Instead, we achieve uniqueness by precisely defining the concept of a {\em walking direction}. The study of Haglund et al  \cite{Haglund89},  constructs a $\TWD $   complex signal comprising modulus,  orientation and phase, by probing with an interpolated quadrature filter too.  However, it differs at two important points. First we obtain {\em all}  critical parameters determining the filter by Total Least Squares (TLS) fitting. Second, our method does not need local spectrum sampling, or their uniform covering of the spectrum. The latter has only discrete solutions for dimensions $ N>2 $.

That is analytic signal without orientation bias  becomes  a formidable task already in $\THD $, as the number of filters  explode in discrete steps if bandwidths need to be varied. 
By contrast, our method must not (but can) use Quadrature filter responses.
Similar to their approach,  our phase is $2\pi$-periodic, and directly relates to the Fourier Transform (FT) phase, as it is obtained by probing it,  and is referred to here as the FT phase. We treat $f$ as a local rather than a global image, a similarity shared with nearly all studies of analytic signals.

An alternative generalization of the $\OND $ analytic signal to $\ND$ uses the Riesz transform,\footnote{The gradient of a low-pass–filtered input, with transfer function $1/|\bom|$.}, strictly assuming the local image being intrinsically $\OND$.
Most studied in $\TWD$, it embeds local phase, scale, direction, and amplitude into a $\THD$ hypercomplex field~\cite{delanghe1978hypercomplex}, known as the monogenic signal—a term unrelated to its  usage in geology or biology.
Its components can be obtained in the Fourier domain or by convolution with the Riesz kernel, using complex~\cite{larkin2001natural}, quaternion~\cite{Felsberg01}, or algebraic geometry~\cite{Felsberg02phd} formulations.

Fundamentally, Monogenic signal phase differs from the FT phase: it loses meaning when the image is not intrinsically $\OND$.
Its wavelet and multiscale extensions were studied in~\cite{Unser09}, where the Structure Tensor was proposed to stabilize its orientation component, ~\cite{alessandrini13,Puspoki16}.

The Structure Tensor (ST) for $N$-dimensional images was introduced in~\cite{bigun87london} via eigenanalysis of gradients and shown equivalent to complex convolutions and pixelwise complex squaring in $\TWD$ images.
A similar 2D matrix formulation appeared independently in~\cite{kass,forstner87}. It has been shown that ST formalism can be extended to fit multiple orientations in $\ND $, or can fit orientation to curvilinear coordinates of  Harmonic functions in $\TWD $.  It enables direction interpolation~\cite{knutsson89tensor}, improving on earlier 5-D mapping of quadrature filter responses~\cite{knutsson1985producing} designed for $\THD $ signals. ST can thus be implemented with   sampled (local) power spectrum by  magnitude responses of quadrature or Gabor filters, beside by responses of Gaussian derivative filters. 

It has since been central to several fundamental aspects of low-level vision, in Gaussian scale space \cite{Laptev05,Lindeberg15}, wavelets \cite{Puspoki16,Bou24},  anisotropic diffusion~\cite{perona1990scale,weickert99,bruhn2005}, corner detection \cite{harris88}, motion estimation \cite{jaehne93,barron94}, and  texture description \cite{rao90,donahue}. It also relates to deep neural networks such as  the Independent Subspace Network~\cite{Hyvarinen09,Le11} and harmonic network, \cite{worrall17}.

         {
  \parbox {\columnwidth} {
In this work, both the ST and Gabor filters are used to reconstruct the analytic signal from local images $f$.
The tensor is used because it provides a Total Least Squares (TLS) estimate not only of local orientation but also, as will be shown, of scale, yielding a complex Gabor filter tuned to the intrinsic $\OND$ content of the image across all components of the planar wave characterizing $f$.
Using scalar products between local images and their adaptively tuned probing filters recovers the missing magnitude and phase components, thereby completing the analytic signal.
}
}

                  {
  \parbox {\columnwidth} {
We provide an implementation (to be released) and quantitative comparisons on datasets with known ground truth, showing superior performance over state-of-the-art methods.
Our approach achieves rotation and scale equivariance of the frequency vector, translation equivariance of the phase, and rotation- and scale-invariance of the TLS errors (eigenvalues) and signal amplitudes.
}}

\section{
Scalar products with Gabor filters \label{sc:analytic_gabor} } 

A Gabor filter with $\br,\bom_0 \in \mathbb{R}^N$ is conveniently defined as
\begin{equation}
g_{\sigma_0^2,\bom_0}(\br)
=\frac{1}{c}
\exp(i\bom_0^\T\!\br)
\exp\!\left(-\frac{\|\br\|^2}{2\sigma_0^2}\right),
\label{eq:gabordcmp}
\end{equation}
where $\sigma_0$ controls its spatial extent and 
$c = (2\pi\sigma_0^2)^{N/2}$ normalizes its volume, 
$\int |g_{\sigma_0^2,\bom_0}|\,d\br = 1$.  
Its frequency transfer function is
\[
G_{\sigma^2,\bom_0}(\bom)
=\exp\!\left(-\frac{\|\bom-\bom_0\|^2}{2\sigma_0^{-2}}\right),
\]
i.e., a Gaussian with inverted variance and unit maximum 
$G_{\sigma^2,\bom_0}(\bom_0)=1$.  
The frequency vector $\bom_0$ can vary {\em arbitrarily}, and we denote by 
${\cal B} = \{ g_{\sigma^2,\bom_k} \}_k$ 
a finite set of such filters.  
In general, a distinct set may be defined for each local image region, 
${\cal B}(\br)$.

By the Parseval–Plancherel theorem, 
the scalar product between $g_{\sigma^2,\bom_k}$ and the local image $f$ 
is preserved between spatial and spectral domains:
\begin{equation}
z_{\sigma_j^2,\bom_j}(\br)
=\langle f, g_{\sigma_j^2,\bom_j}\rangle
=\langle F, G_{\sigma_j^2,\bom_j}\rangle.
\end{equation}
If ${\cal B}(\br)$ is chosen such that its members are tuned to the dominant frequencies $\bom_k$ of the local image $f$, 
and if these $\bom_k$ lie on the same side of an arbitrary hyperplane through the spectral origin with normal $\bn$, 
then the sum
\begin{equation}
f_A(\br)=\sum_{\bom_j\in {\cal B}(\br)} z_{\sigma_j^2,\bom_j}(\br)
\end{equation}
approximates the analytic signal increasingly well as the 
cardinality of ${\cal B}$ grows.
However, for now we assume that the cardinality of all ${\cal B}(\br)$ is one,
so that
\begin{equation}
f_A(\br) = z_{\sigma_j^2,\bom_j}(\br). \label{eq:onewave_assumption} 
\end{equation}
Although this condition is generally not known to hold, adopting it allows us to
perform image enhancement by achieving non-linear, isotropic smoothing along
local orientations while preserving analytic singularities, as demonstrated in
the experimental results.

{
  \parbox{\columnwidth}{
Each $g_{\sigma^2, \bom_j}$ has a distinct {\em filter walking
direction} defined by $\bom_k / \|\bom_k\|$. 
It specifies the direction in which one ``walks'' in the image 
to achieve maximal phase increase upon filter application.  
All filter walking directions lie within the same hemisphere, 
referred to as the {\em mean walking direction}, determined by $\bn$.
  }
}

\section{Structure Tensor driven Phase  \label{sc:stdphase}}

Two Structure Tensors (STs) are used in this work: one optimizing the local direction,
denoted $\bS^D(\br)$, and another optimizing the local scale,
denoted $\bS^S(\br)$. Both are computed for the local image around an
image point $\br \in \mathbb{R}^N$.

The directional tensor $\bS^D$ is the well-known $N \times N$ matrix
\begin{equation}
\bS^D(\br) = \langle \nabla f, \nabla^\T f \rangle.
\end{equation}
The tensor $\bS^D$ is a non-linear transform obtained for local images
by chaining (linear) convolutions, (non-linear) pointwise
multiplications, and (linear) convolutions in sequence.
The most significant eigenvector of $\bS^D$ fits a TLS-optimal line
passing through the origin of $F$, with eigenvalues describing modeling
errors \cite{bigun87london}. These are used in various combinations to
measure applicability, coherence, reliability, cornerness, and other
properties. 

Accurate absolute frequency (also known as scale) is a valuable texture
descriptor. However, estimating it reliably and densely is challenging,
especially when the directional composition of the local image is
unknown, as is the case here. To alleviate this problem, we take a TLS
approach. The method, summarized below, is based on modeling the trace
of the ST, $\Tr(\bS^D)$, under changes of the inner scale (which we can
control) of its gradient computations \cite{bigun16lasvegas}.

The local image $f$ is approximated by a set of sinusoids:
\begin{equation}
\hat f(\boldsymbol r)
  = \sum_{\bom_k \in {\cal B}(\br)} A_k \cos(\bom_k^\T \br + \varphi_k),
  \label{eq:csin}
\end{equation}
where $A_k$, $\bom_k$, $\varphi_k$, and the cardinality of ${\cal B}$
are unknown. However, it is assumed\footnote{The assumption can be
enforced by bandpass filtering if violated.} that the local image has
one dominant scale, and that all frequency vectors $\bom_k$, while
having different directions as far as scale estimation is concerned, share the same magnitude
$\omega_0 = \|\bom_k\| > 0$.

A new trace $\Tr(\bS^D(\br))$,
\begin{equation}
\Tr(\bS^D_{\sigma^2_{\text{in}}})(\br)
  = \langle \nabla^\T_{\sigma^2_{\text{in}}} f,
            \nabla_{\sigma^2_{\text{in}}} f \rangle,
\end{equation}
can be computed by varying $\sigma^2_{\text{in}}$, and these traces are
in general not identical. In this expression, the gradient of a Gaussian
with variance $\sigma^2_{\text{in}}$ is spatially sampled and used in
gradient filtering. This discrete and separable gradient filter
convolves the full-sized image to produce sampled gradients.
Consequently, the gradient and the tensor are denoted
$\nabla_{\sigma^2_{\text{in}}}$ and $\bS^D_{\sigma^2_{\text{in}}}$,
respectively, to emphasize the dependency.  
Although not explicit in the equation, there is also an outer scale
$\sigma^2_{\text{out}}$ defining the neighborhood size, i.e., the
support of the scalar product $\langle \cdot,\cdot\rangle$.
The outer scale is chosen in fixed proportion to the inner scale, so it
introduces no additional freedom.

The \emph{logarithm of the Structure Tensor trace} defines a
(non-linear) scale space in which a local image with a well-defined
scale—possibly containing multiple directions—generates a straight line:
\begin{equation}
\log \Tr(\bS^D_{\sigma^2_{\text{in}}}(\br))
  = C - \omega_0^2 \sigma^2_{\text{in}},
  \quad \br \in \mathbb{R}^N,
  \label{eq:li11_smai}
\end{equation}
whose direction is given by the tangent $-\omega_0^2$. The constant $C$
is independent of the inner scale $\sigma^2_{\text{in}}$ used in
$\nabla_{\sigma^2_{\text{in}}}$.

Scale can therefore be estimated by computing
$\log \Tr(\bS^D_{\sigma^2_{\text{in}}})$ at several predetermined inner
scales $\sigma^2_{\text{in}}$ for all $\br$, and then fitting a
direction to the observation points
$(\sigma^2_{\text{in}},\, \log \Tr(\bS^D_{\sigma^2_{\text{in}}}))^\T$.
These points  are always two-dimensional
(inner scale vs. trace), independent of the image dimension $N$.

Thus, TLS line-direction fitting to
(\ref{eq:li11_smai}) is carried out by a \emph{two-dimensional} (scale-estimating)
Structure Tensor, denoted $\bS^S$, not to be confused with $\bS^D$.
The required 2D gradient vectors of this tensor are obtained by applying
finite differences to \emph{consecutive} observation points as the inner
scale increases. All such vectors form one scale neighborhood at each
$\br$ and must agree in direction to produce a low TLS error. Hence, the
TLS direction fitting is performed through the eigenanalysis of $\bS^S$
in the log-scale space determined by inner-scale sampling of $\bS^D$.

With the vector $\bom(f)$ that best explains the local image $f$ at hand—its direction and magnitude determined by the eigenvectors of $\bS^D$ and $\bS^S$, respectively—we compute the \emph{Structure Tensor phase}, denoted $\Phi^{\bST}$. It is obtained by taking the scalar product between the local image and a Gabor filter tuned to that image, whose frequency vector is $\bom(f)$ and whose variance $\sigma^2$ is directly determined by $\|\bom(f)\|$, as $\sigma(\bom(f)) = \alpha \|\bom(f)\|$ with $\alpha$ being a fixed constant. The phase and amplitude are then given by the argument and magnitude of
scalar product, where the gabor filter changes with the local image $ f $. 
\begin{equation}
\begin{aligned}
z_{\sigma^2,\bom(f)}(\br)
   &= \langle f, (g_{\sigma^2(\bom(f)),\bom(f)} )\rangle \\
   &= u(\br) + i v(\br)
    = m(\br)\exp(i\Phi^{\bST}(\br)).
\end{aligned}
\label{eq:z_stdp}
\end{equation}
Computing such scalar products,  at all $ \br$ yields ST analytic signal, comprising local amplitude and phase,
\begin{equation}
f_A(\br)=z_{\sigma^2,\bom(f)}(\br)=m(\br)\exp(i\Phi^{\bST}(\br))
\end{equation}
as a consequence of (\ref{eq:onewave_assumption}).

Eigenvectors, even when unique, are defined only up to sign, including those
obtained from a Structure Tensor. When an eigenvector is used as an estimate of
$\bom/\|\bom\|$ to parameterize a Gabor filter, its sign must therefore be
resolved. We remove this ambiguity by assigning each eigenvector to a fixed
hemisphere defined by an arbitrary but fixed vector $\bn$, which is the same for
all local images. If $\bom^{\top}\bn < 0$, the vector is replaced by $-\bom$.
This yields a unique spatial direction for every filter response, $\bom$, along
which the phase response increases monotonically and maximally in the image, as
long as the response magnitude is non-zero.

The resulting, sign-corrected direction of $\bom$ is called the \emph{filter
walking direction}. Since all filters are forced into the same hemisphere with
respect to the chosen normal $\bn$ of the separating hyperplane, we refer to
this common hemisphere orientation as the \emph{image walking direction}.
However, sign determination is more reliable when $\bom$ lies close to the
direction of $\bn$ than when it lies farther away; avoiding artifacts caused by
this reduced certainty therefore requires an additional strategy, which is
described in the Supplementary Material.

The dependence of $\Phi^{\bST}$ on a
fixed image walking direction $\bn$ is crucial for obtaining a phase that
exhibits only $2\pi$ discontinuities\footnote{Jumps of exactly $2\pi$ are not
artifacts but inherent to phase, since the observed phase of a real image also serve as the
argument of a complex sinusoid.}. 
When this condition is met, $\nabla \Phi^{\bST}$ is continuous everywhere in the image including at $ 2\pi$ jumps, i.e. in both magnitude
and direction. The only exception to directional continuity occurs in local
images whose frequency vector is orthogonal to the separator normal, i.e.\ when
$\bom^{\top}\bn = 0$. Details  on how to perform $\nabla \Phi^{\bST} $ to have the desired
continuity is given in the Supplementary Material.

We compute the analytic signal 
 where it is meaningful, i.e., where the eigenvalues of ST reliably
 indicate that $f$ contains a dominant direction and scale. Else, we
 put the signal value to zero.


\section{Experiments}
\subsection{Test images with known ground truth \label{sc:minutia_model}}

Artificial bias with respect to orientation, scale, and singularities can be
assessed only if ground truths for these properties are available, and ideally
if these properties are also uniformly distributed in the test inputs. Most
natural images, and most general-purpose test sets, fail to satisfy one or more
of these criteria. If a method is unbiased with respect to orientation and
scale, then it is also unbiased with respect to analytic phase, since the
derivative of the analytic phase is the frequency vector—the embodiment of
local orientation and scale.

We use two families of test images: one without analytic
singularities, and one containing 70 analytic singularities in every test
image, with known ground truth for orientation, scale, phase, amplitude, and
singularity location. Both families represent meaningful \emph{local} models of
natural images, although not necessarily global ones. The second family is
especially relevant for testing behavior near corners, bifurcations, and
line-endings, in particular fingerprint singularities.

\begin{figure}[t]
\begin{minipage}[t][][t]{\columnwidth}
\fbox{
\includegraphics[width=0.96\columnwidth]{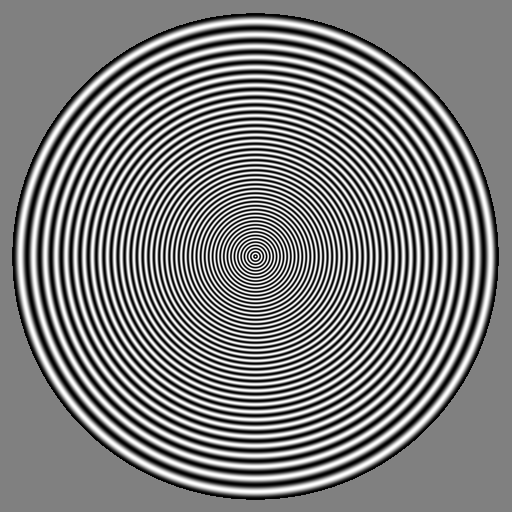}%
}
\caption{\label{fg:fmtest_65_35.png}  {\bf
FM test image} with size 512x512.  The period changes linearly with
  radius.}
\end{minipage}
\end{figure}

\noindent\small\textbf{Frequency-modulated (FM) test images}\par
\vspace{0.25em}
\noindent
 Fig. \ref{fg:fmtest_65_35.png} represents a clean  sinusoid  varying slowly in
 frequency vector, where the phase is changing  logarithmically
\begin{equation}
\cos (C\,\log\|\br\|),
\label{eq:cos_phi_L_our}
\end{equation}
The gradient of the phase, which equals to the instantaneous frequency
vector,   is
then given by 
\begin{equation}
\nabla (C \log r) = \frac{C}{r} \mathbf{e}_r
\end{equation} 

The frequency vector is 
agnostic to direction, i.e. isotropic, angular wise. Its  magnitude
is the absolute frequency,  $ \omega =C/r $, the scale. Thus,  the
instantaneous wave period is $ T= \frac {2\pi }
{C} r $ and varries
perfectly linearly, which is important as a ground-truth for
evaluation purposes.

\begin{figure}[t]
    \centering
\begin{minipage}[t][][t]{\Limsize}
  \fbox{
  \includegraphics[width=0.96\columnwidth]{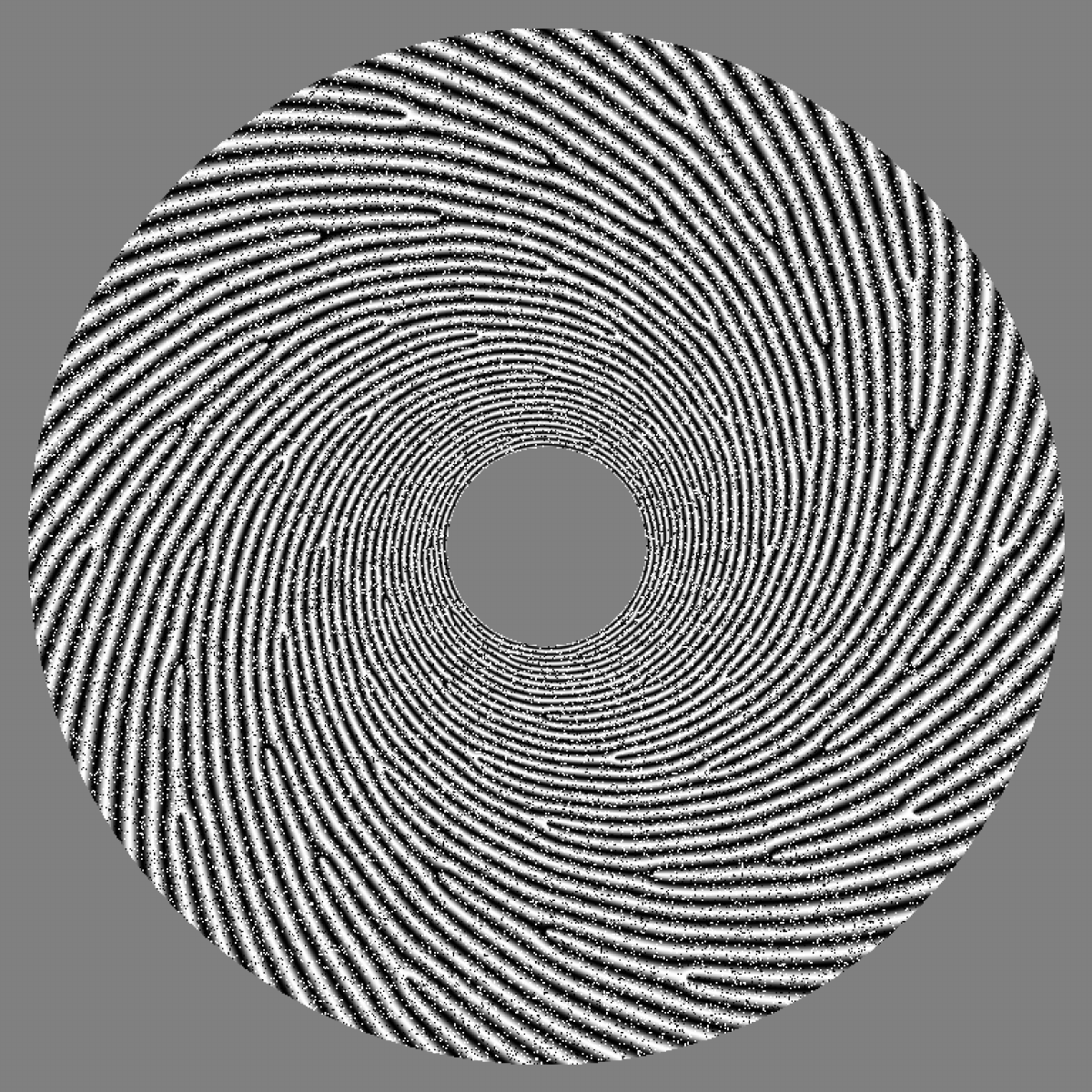}
}
  \caption{\label{fg:minutia_synt_}\label{fg:minutia_synt_mm_sp_noise_infull} 
{\bf Singularities}  are produced by phase gradient superpositions in
image with size 758x768.
Contamination is replacement noise, i.e. not additive. Al ground-truths of singularity parameters, as well as
those of the phase, are known. 
}%
\end{minipage}
\end{figure}

\begin{figure}[t]
\begin{minipage}[t][][t]{\columnwidth}
\fbox{
\includegraphics[width=0.96\columnwidth]{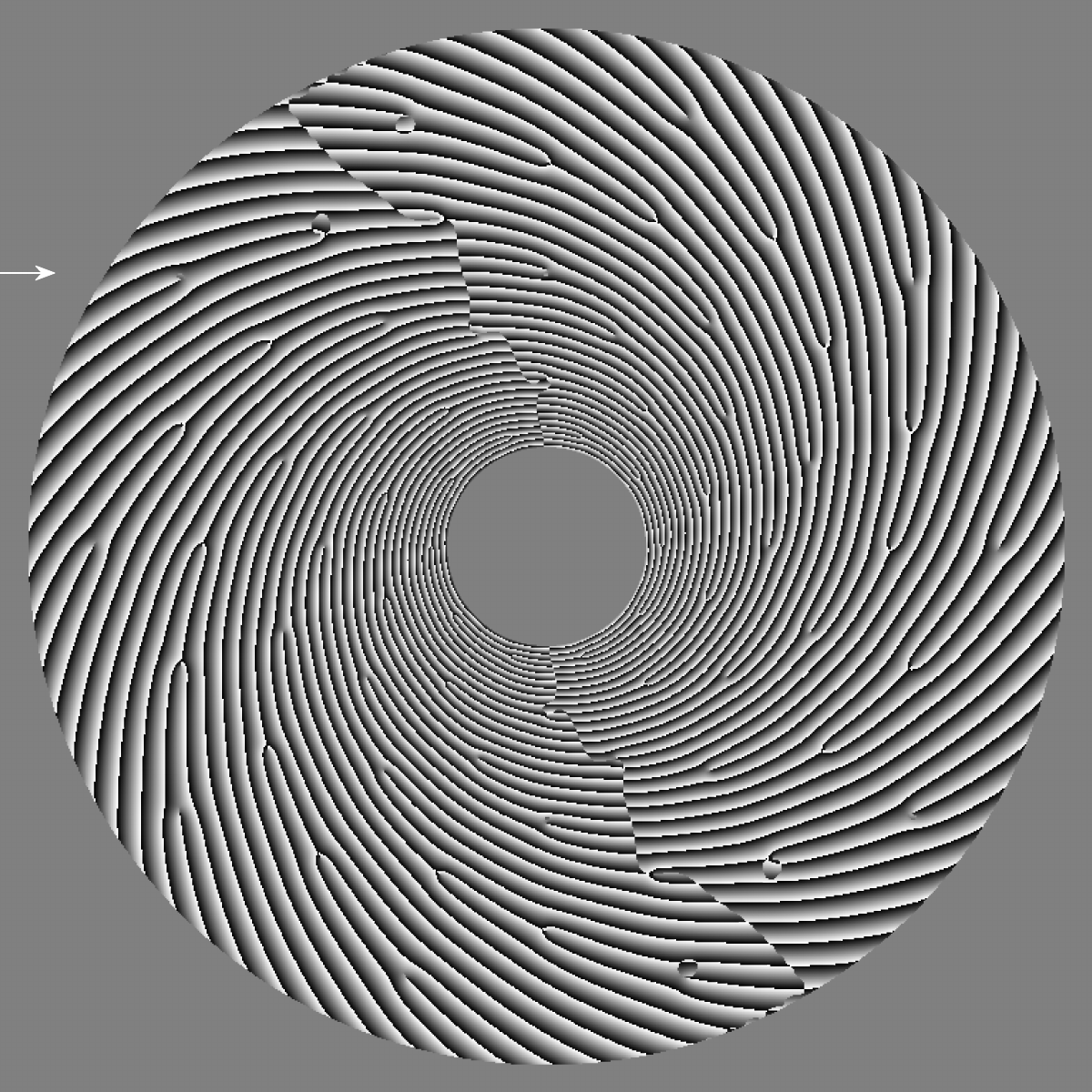}%
}
\caption{\label{fg:minutia_synt_lstd}%
{\bf ST phase} of clean input (underlying  Fig. \ref{fg:minutia_synt_mm_sp_noise_infull});
}%
\end{minipage}
\end{figure} 

\noindent\small\textbf{Test images with singularities}\par
\noindent
We summarize  the analytic model used to generate the synthetic
images, exemplified 
by Fig.~\ref{fg:minutia_synt_}, together with the corresponding ground
truth for local image properties: orientation, scale, phase, amplitude,
location, singularity presence, and associated singularity parameters. The formulation combines a linear phase
field $\Phi_L$ with localized phase perturbations $\Phi_P$, inspired by
fingerprint minutia models \cite{sherlock93, cappelli07, cao15} and by fringe
pattern models in waves and optics \cite{larkin07synth}.

Embedding $J$ singularities at locations $\br_j$,  neighborhoods of which are
well approximated by planar waves, with frequency vectors $\bom_j$ were it not for the embedding of respective singularities,
(at each $\br_j$)  yields
\begin{equation}
\Phi(\br)
= \Phi_L(\br)
+ \sum_{j=1}^{J} (-1)^{t_j}
    \arg\!\left(\frac{\tilde\br - \tilde\br_j}{i\tilde\bom_j}\right).
\label{eq:combined_minutia_model_main}
\end{equation}
\emph{The operator $\tilde{\cdot}$ } maps 2-D vectors to their complex-number
representation, and $t_j\in\{0,1\}$ specifies the singularity type, whereas $ \bom_j$ are computed
analyticaly.

Once the singularity locations and the slowly varying background phase
$\Phi_L(\br)$ are specified—where $\Phi_L(\br)$ is locally well
approximated by the phase of a planar wave, namely $\bom_0^\T\br$—the
superposition of phases in (\ref{eq:combined_minutia_model_main}) determines
the complete analytic phase. The associated analytic signal of the test image is then
\[
f_A(\br) = \exp(i\,\Phi(\br)),
\]
where we set the amplitude to $1$. The clean test image is obtained as $\Re{f_A}$.
We generate numerous such images by changing parameters, and degrade
them by noise  for evaluations.

We contaminate the clean image non-linearly using Salt\&Pepper Replacement (SPR)
noise at a controllable SNR\footnote{Signal-to-Noise Ratio for SPR noise is $1-({\#\mathrm{contaminators}/\#\mathrm{pixels}})^{1/N} $, i.e., the
average fraction of clean pixels in any direction in the image.}. In the figure, the SNR is~0.61.
The contamination is applied only after the clean test image has been
synthesized.

In  Fig.  \ref{fg:minutia_synt_mm_sp_noise_infull} we  have 7 sets of singularities,
each set with a common 
 ridge  period, which increases linearly by increment
 of $ \approx$ 2 pixels 
 as distance to 
image centre grows.  There are 10 singularities with
different directions but same singularity type in  each set  and the   type alternates between sets. 
The image simulates thereby both scale  and pattern variations    of singularities.  The ground truth, concerning location, direction,
type and period of singularities are known by construction.

\begin{figure}[t]
    \centering
\begin{minipage}[t][][t]{\Limsize}
  \fbox{
  \includegraphics[width=0.96\columnwidth]{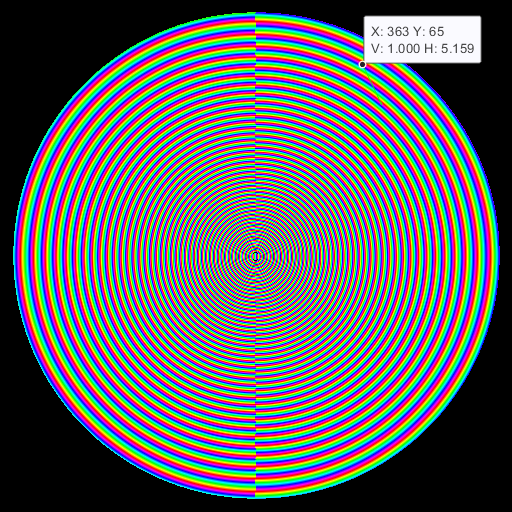}
}
  \caption{\label{fg:st_sa_fmtest_65_35} 
{\bf ST} complex signal. Hue represents signal argument and brighntess
is its magnitude, in HSV colors.
}%
\end{minipage}
\end{figure}

\begin{figure}[t]
    \centering
\begin{minipage}[t][][t]{\Limsize}
  \fbox{
  \includegraphics[width=0.96\columnwidth]{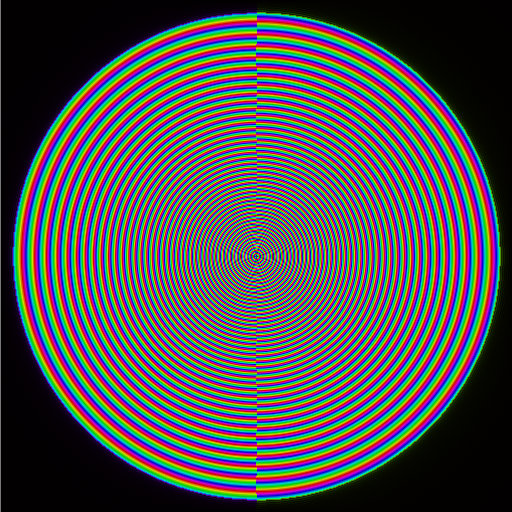}
}
  \caption{\label{fg:MP_MP_fmtest_65_35} 
{\bf Monogenic} complex signal. Hue represents signal argument and brighntess
is its magnitude, in HSV colors.
}%
\end{minipage}
\end{figure}

\subsection{\label{sc:mod_bias} Isotropy of  modulus and Phase} 

 Figures \ref{fg:st_sa_fmtest_65_35}  and
 \ref{fg:MP_MP_fmtest_65_35} show our ST  Analytic signal and Monogenic Analytic
 Signal (MS), \cite{larkin2001natural}\cite{Felsberg01} 
 where pixels  are similarly represented. Colors are in HSV
 representation, where brightess represents the modulus of analytic
 signal whereas hue represents its argument. The saturation is set to
 maximum.  In valid signal area both signals have magnitudes which are
  constant (ST signal is $ 1.000 $,  MS signal is $1.00\pm 0.01 $ in
  the applicable image area).

  The phase of ST and MS vary linearly and without bias. The phase linearity
  is in the ST case in $[-pi,\pi]$ whereas in MS case it is
  $[-pi/2,\pi/2]$. Note  that  Fig. \ref{fg:MP_MP_fmtest_65_35} uses the
  full hue gammut (not the half).

 Thus, from clean images results of ST signal concerning orientation
 bias, and the representation of amplitude we conclude that ST
 analytic signal behaves well. In that respect it agrees with  the
 well known MS signal, except the phase range.

\begin{figure}[t]
\begin{minipage}[t][][t]{\columnwidth}
\fbox{
\includegraphics[width=0.96\columnwidth]{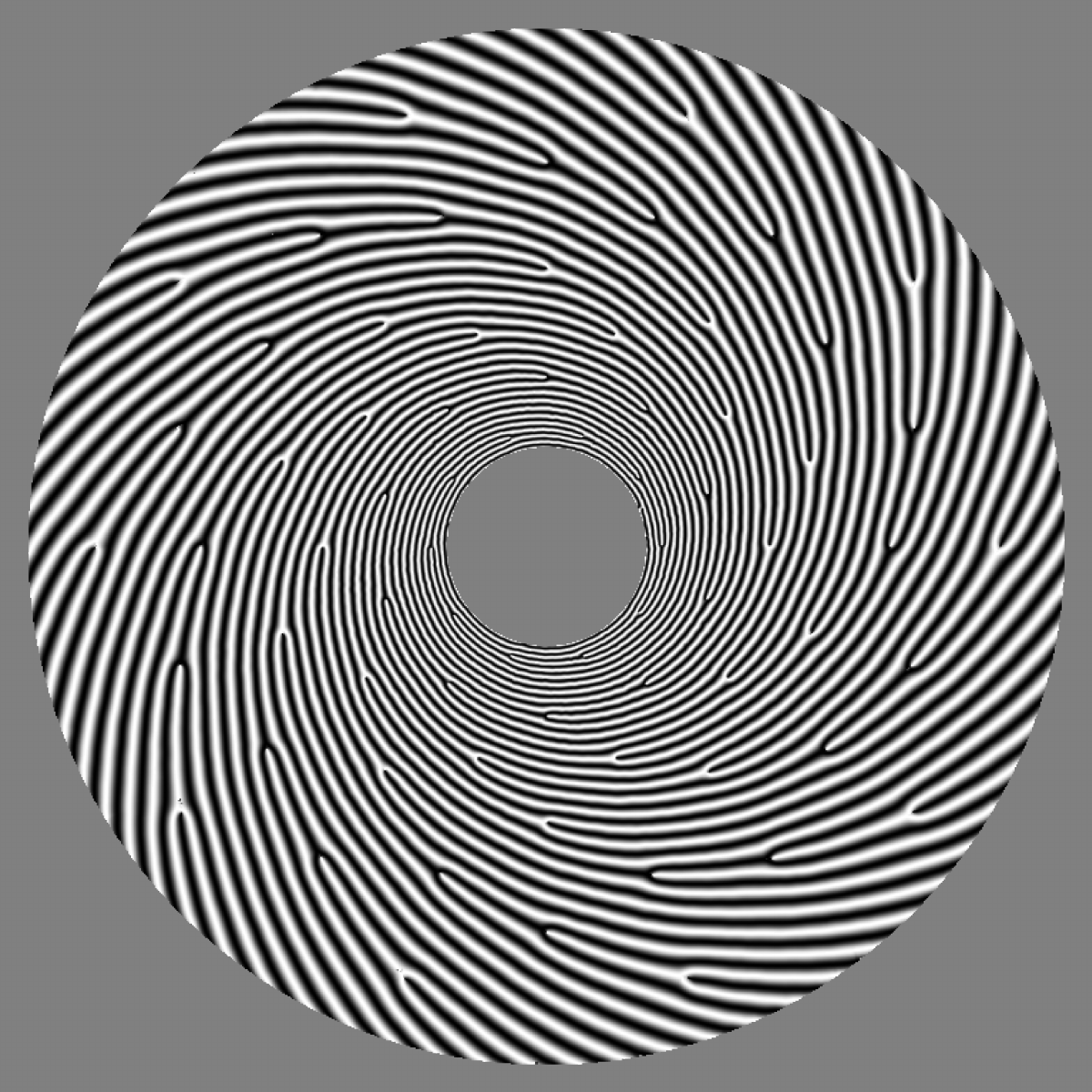}%
}
\caption{\label{fg:minutia_synt_mm_sp_noise_infull_reconst}  {\bf
    Reconstructed image} from the ST phase of
  Fig. \ref{fg:minutia_synt_mm_sp_noise_infull}. }
\end{minipage}
\end{figure} 
 
\begin{figure}[t]
\begin{minipage}[t][][t]{\columnwidth}
  \fbox{%
  \includegraphics[width=0.96\columnwidth]{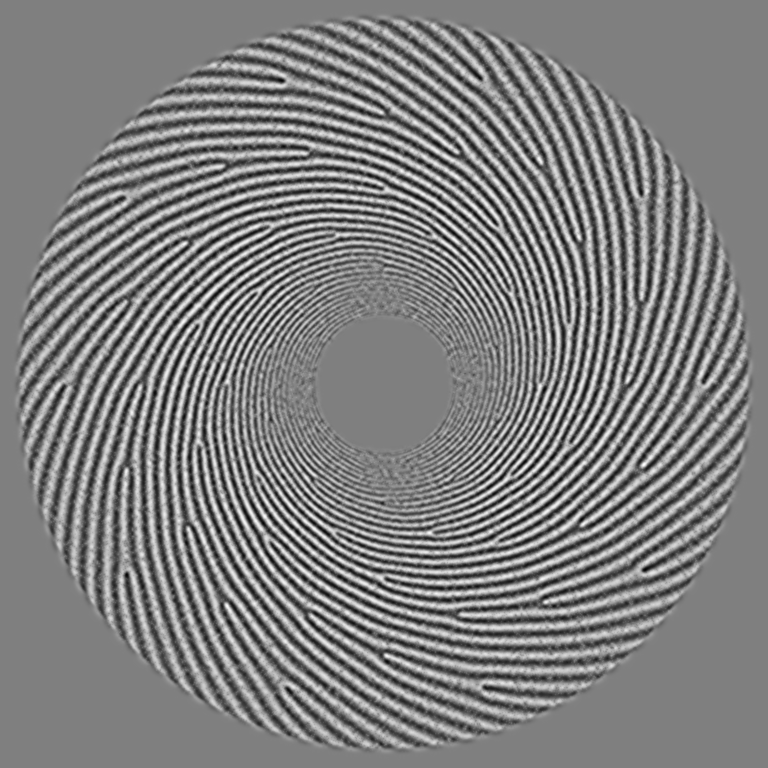}
} 
  \caption{\label{fg:RE_MS_minutia_sp_noise_infull}
  {\bf Reconstructed image} from monogenic phase applied to 3rd level
  (of total 5), a Laplacian like
  pyramid, \cite{Unser09}. }
\end{minipage}
\end{figure}

\subsection{\label{sc:artifact} Enhancement:  singularities \& contamination}

Figure~\ref{fg:RE_MS_minutia_sp_noise_infull} shows the reconstructed
image obtained from the ST phase of the noisy input
(Fig.~\ref{fg:minutia_synt_mm_sp_noise_infull}) via
$\Re{\exp(i\,\Phi^{\bST})}$. As can be seen from this result—and
consistently across many test images over a 
wide range of scales—  
(i) the reconstruction is nearly flawless\footnote{A reader with acute
vision  can verify that
there is a tiny noise, of the massive noise presence of the original image,  nonetheless has
filtered through (Bottom left of
 the ``circle''),  as  tiny deformations at the (inner) border.}, not only for ridge patterns (which
are locally $\OND$) but also for singularities, despite the non-linear
contamination;  
(ii) the mapping $\Re{\exp(i\,\Phi^{\bST})}$ is invariant to both $\pm\bom$ and
$2\pi$ phase wraps, explaining the disappearance of these artifacts;   
(iii) the ST phase is isotropic and remains stable across a large span of
scales with fidelity to ground-truth  (constant)  signal amplitude.

Figure~\ref{fg:minutia_synt_lstd} shows the ST phase computed from a noise-free
image, highlighting that discontinuities are inherent to $\Phi^{\bST}$ even when
the data contain no noise, yet $\Re{\exp(i\,\Phi^{\bST})}$ remains continuous.
These discontinuities arise because $\Phi^{\bST}$ must be reconstructed from
observations of the real part of $f_A$ alone when forming the ana lytic signal.
Two types of artifacts appear: {\em circular discontinuities} and {\em
hyperplane discontinuities}.

Circular discontinuities correspond to familiar phase-unwrapping jumps—i.e., discontinuities of $2\pi $  arising from the periodic nature of the reconstructed phase, which contrasts with the truly continuous ground-truth phase (illustrated in Supplementary Material).
 In Fig.~\ref{fg:minutia_synt_lstd}, they appear as radial, sawtooth-shaped gaps. Between these gaps, the phase varies linearly in the radial direction—showing a sawtooth profile when zoomed in—not to be confused with the sinusoidal variation of grayvalues of the original signal.

Hyperplane discontinuities, on the other hand, appear near the vertical midline
on horizontal wavefronts. They occur at locations where the local wave vector
$\bom$ satisfies $\bom^{\top}\bn=0$. In such regions, small changes in $\br$ may
cause $\bom$ to cross the zero-separation boundary in the Fourier domain,
producing sign flips. These jumps (a) occur only in $\ND$ with $N>1$, (b) have
phase-gaps of  $2|\bom^{\top}\br|$ (instead of $2\pi $), and (c) are invisible in observations where
$\pm\bom$ are equivalent (e.g.\ in $\Re{f_A}$), but remain visible in
representations that expose phase directly (e.g.\ $\arg f_A$).

As a comparison, we attempted to reconstruct the clean image from the contaminated one using the phase of the Monogenic Signal.
Taking its real part, as with ST-phase, reproduces the contaminated image by construction; thus, alternative strategies are needed for enhancement. Recent work suggests combining the Monogenic Signal with ST or Laplacian-like pyramids. We explored a spline-wavelet pyramid guided by ST \cite{Unser09}, applying the Monogenic Signal at each level to obtain the imaginary part of the analytic signal. The best result, shown in Fig.~\ref{fg:RE_MS_minutia_sp_noise_infull}, corresponds to one manually selected Laplacian pyramid level using the public implementation by \cite{Unser09,Puspoki16}, without down-sampling between levels.

The observed non-isotropy arises from the spline-wavelet pyramid, which is inherently non-isotropic at high frequencies, whereas the Riesz transform used for the imaginary part of the Monogenic Signal is isotropic. Manual selection excluded images that were (i) noisier and (ii) even less isotropic than the figure.

In summary, our analytic signal  $f_A$, 
is a continuous, isotropic approximation of the analytic signal, capable of capturing scale variations while applying directional smoothing that respects analytic singularities, and iso-curves. It enhances structures compatible with the underlying single-orientation, single-scale model, while deemphasizing incompatible ones—typically noise, as observable in our results.  If the suppressed content is not noise but meaningful structure, it must be addressed by extending limits of the assumption.

\begin{figure}[t]
\begin{minipage}[t][][t]{\columnwidth}
\fbox{%
\includegraphics[width=0.96\columnwidth]{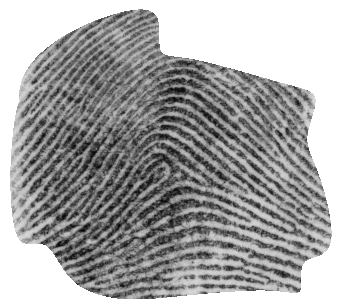}%
}
\caption{\label{fg:112_0_im_valid_cut} {\bf Fingerprint
image} of size   308 x342 pixels}
\end{minipage}
\end{figure}

\begin{figure}[t]
\begin{minipage}[t][][t]{\columnwidth}
\fbox{%
\includegraphics[width=0.96\columnwidth]{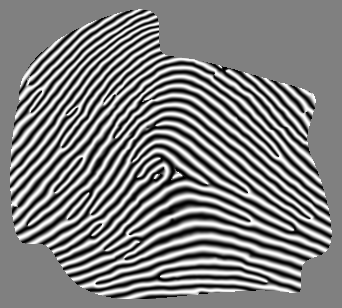}%
}
\caption{\label{fg:st_sa_real_sd27_112_01_2025-11-21} {\bf Enhanced  Fingerprint}}
\end{minipage}
\end{figure}

 \subsection{Enhancement: Fingerprints}

In Fig. \ref{fg:112_0_im_valid_cut}, a fingerprint is shown. We
enhanced fingerprints using our ST-phase method, as in Sec. \ref{sc:mod_bias}
and \ref{sc:artifact}, except that the phase amplitude is
normalized. The result is shown in
Fig. \ref{fg:st_sa_real_sd27_112_01_2025-11-21}. As can be verified
visually all minutia and ridge
structures are kept in tact, while clarity has improved. This and
similar results indicate that incorporating the real part of the ST
phase improves ridge clarity and continuity as well as minutia, beyond what current state-of-the-art methods achieve.

\begin{figure}[t]
\begin{minipage}[t][][t]{\columnwidth}
  \fbox{%
  \includegraphics[width=0.96\columnwidth]{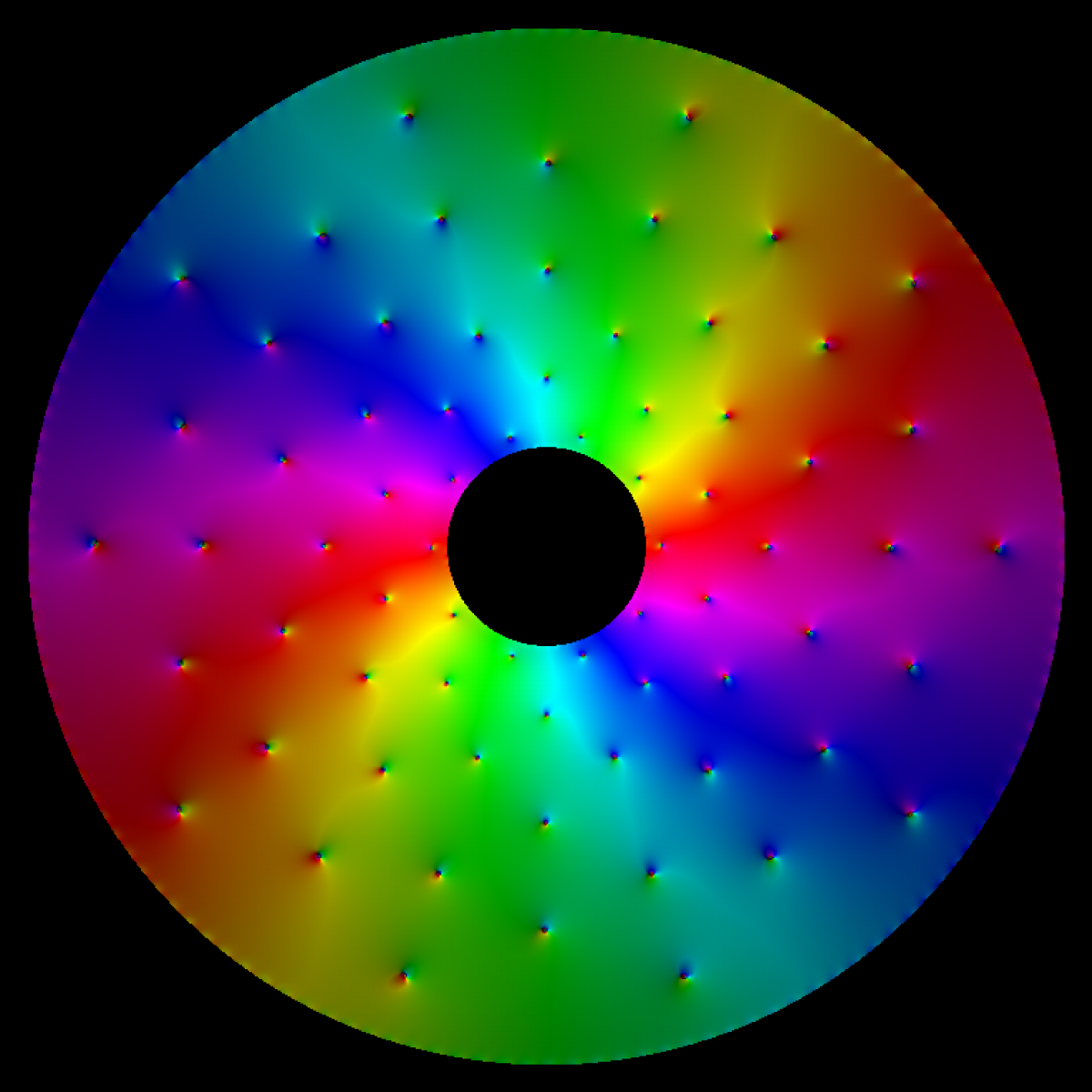}%
  }
  \caption{\label{fg:phsr_Daleth_COMP_omest}
  {\bf Complex image} with pixels representing $(\tilde \nabla
  \Phi^{\bST})^2$, eliminating hyperplane artifacts, i.e. both
  magnitudes and arguments are continuous. Hue and intensity depict
  angles and magnitudes of complex pixels, in HSV color space.}
\end{minipage}
\end{figure}

\subsection{\label{sc:true} Detecting analytic singularities in 2-D}
Singularities often indicate abrupt structural changes. We target detection of line endings and bifurcations, previously modeled analytically. Applications include fingerprint minutiae and optical fringe patterns, where accurate detection is critical.

Using $\nabla \Phi^{\bST}$ is central to our approach. Phase gradients have been used for SAR phase unwrapping \cite{ghiglia1998two}, but existing methods are vulnerable to separator-plane singularities. Applying an $\ND$ gradient to real and imaginary parts—rather than phase \cite{Fleet90}—is an elegant alternative. However, artifacts near the separator are revealed in Sec.~\ref{sc:gradph}, where mitigation strategies are also detailed.

\begin{figure}[t]
\begin{minipage}[t][][t]{\columnwidth}
\fbox{\includegraphics[width=0.96\columnwidth]{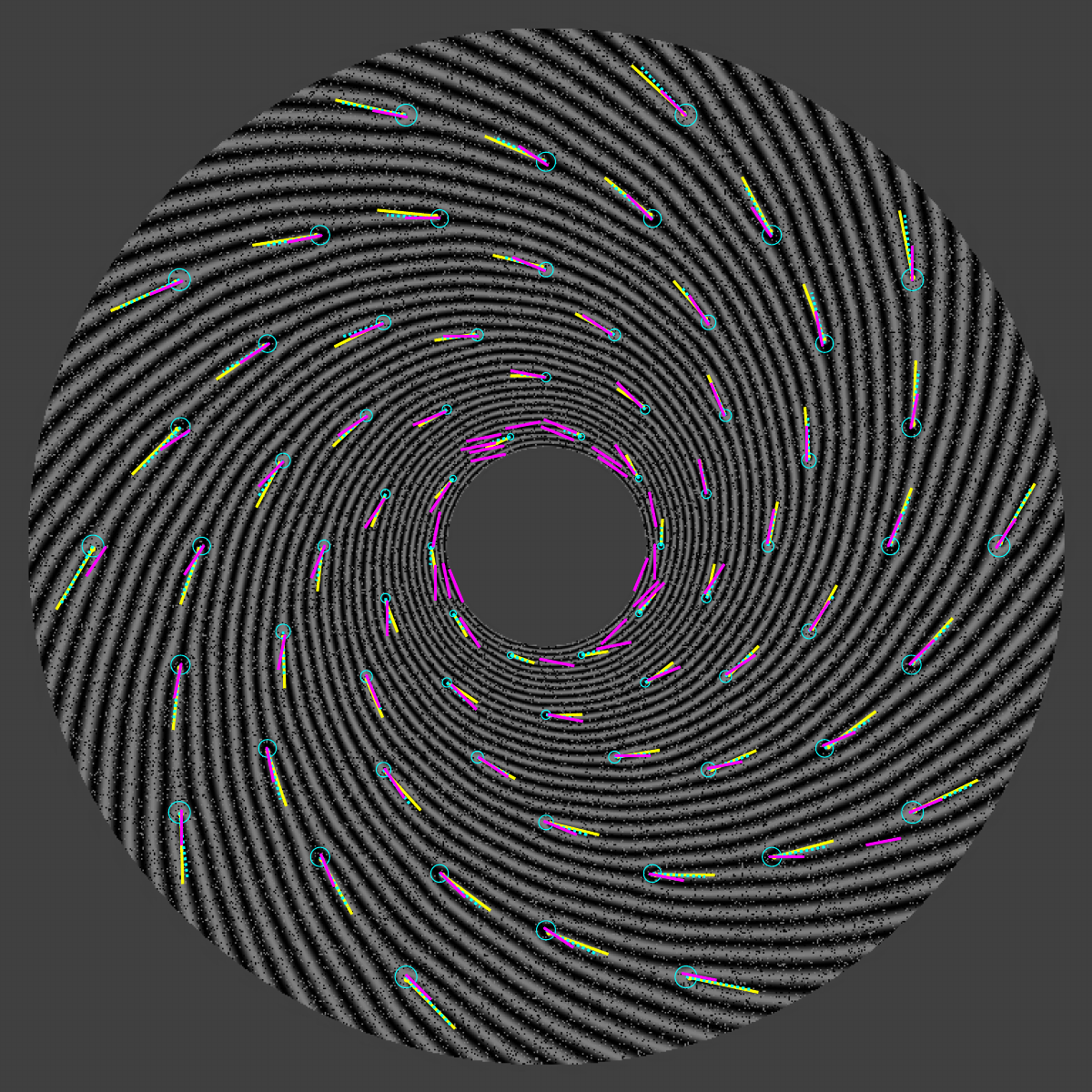}}
\caption{\label{fg:minutia_synt_mm_sp_noise_infull_agr} {\bf Detected singularities} aggregated across scales (yellow), baseline {\bf mindtct} \cite{watson2007user} (pink), and ground truth (cyan).}
\end{minipage}
\end{figure}

We detect singularities modeled in Sec.~\ref{sc:minutia_model} by computing the phase gradient and squaring it as a complex number via $\tilde{\cdot}$:
\begin{equation}
(\tilde{\nabla} \Phi^{\bST})^2 = \tilde{\bom}^2 + \frac{2i\tilde{\bom}}{r}\exp(i\varphi) - \frac{1}{r^2}\exp(i2\varphi)
\label{eq:cmpstphase}
\end{equation}
Here, complex squaring yields the infinitesimal ST. Fig.~\ref{fg:phsr_Daleth_COMP_omest} shows the result, free of separator artifacts.

Projecting this expression onto a filter with phase term $w =
i\exp(i\varphi)$ annihilates the first and third components (since
$\exp(i\varphi)$ is orthogonal to $\exp(i2\varphi)$), resulting in a
high-magnitude complex value aligned with $\tilde{\bom}$ at the
singularity location\footnote{The singularity type can be obtained by
projecting onto $w^*$.}. The pixel magnitude indicates the certainty
that the location corresponds to a singularity, and the argument gives
its direction\footnote{We adopt the ISO standard defining (minutia)
direction,  corresponding to the tangent at the singularity aligning
with ridges, Fig.~\ref{fg:minutia_synt_mm_sp_noise_infull_agr}.}.

The period at singularity neighborhood can be derived from the inverse
of our scale, $|\bom| = |\nabla \Phi^{\bST}|$, which is a novel
feature in this formulation. These computations are demonstrated in
Fig.~\ref{fg:minutia_synt_mm_sp_noise_infull_agr}, where the detected
singularities (in yellow) are overlaid on a dimmed version of the
original image for clarity. Vector magnitudes further represent the estimated singularity periods.

Projecting neighborhoods of $ (\tilde \nabla
\Phi^{\bST})^2 $ 
on $w^2 $, (\ref{eq:cmpstphase}) will  annihilate the
first and the second terms,  retaining the third term of the equation,
extracting the singularity location. 
 The direction information $ \tilde \bom$ vanishes, but the result is
a further evidence for singularity location in noisy images, beside  the projection on $ w$.

Such projections thereby define a  framework for (singularity)
localization, direction, and type determination. Projections are
implemented via complex convolutions, which require angular continuity
of  ST of phase, represented as a complex number (rather than a 2D
matrix). For the continutity we refer to Sec.~\ref{sc:gradph}.

As scale, we  used $\| \bom\|
$ computed to deterimine  probe (Gabor) filters.  This is an analogue of the scale
concept of \cite{Witkin83}, \cite{koenderink90},
\cite{florack1994linear},\cite{lindeberg98} with the main difference
being that ours is obtained by the trace of the ST of the image, 
non-linear with it.

\begin{table}[t]
\begin{minipage}[t][][t]{\columnwidth}
  \begin{tabular}{cc|c}
    \begin{tabular}{l}
   \\
         {\bf present}\\
         {\bf mindtct}\\
\end{tabular}
        &
\begin{tabular}{rr}
  \hline
      {\bf FA}  & {\bf FR} \\
  \hline
     0.0  & 0.0\\
     0.34  & 0.20\\
\end{tabular}
&
   \begin{tabular}{ccc}
    \hline
  { $ \sigma_{loc} $  }  &  $ \sigma_{dir} $   &   $ \sigma_{per} $  \\
  \hline
     0.08  & 0.05 &0.03\\
     0.08  & 0.39 & -\\
\end{tabular}
\end{tabular}
  \caption{\label{tb:fafr} {\bf Singularity}  detection performance with  degraded images.}
\end{minipage}
\end{table}

Table \ref{tb:fafr} compares the detection accuracies of our method and a baseline method, {\bf mindtct} \cite{watson2007user}, when the noise level is increased until the present method starts to deteriorate from 0 \% FR and FA error. This occurred at an SNR of 0.61 (e.g., Fig. \ref{fg:minutia_synt_mm_sp_noise_infull}) when {\bf Salt \& Pepper replacement}, SPR, noise was used as a contaminant. At this point, the baseline operated at 34\% FA and 20\% FR error (pink), whereas our method remained at 0\% FA and 0\% FR. Detection is considered correct if the singularity location lies within half the singularity period of the ground truth. Ground truths are depicted in cyan, including periods shown as vector magnitudes and circles.

Direction estimation errors were within 0.07 radians for our method and 0.14 radians for the baseline. The standard deviation of direction estimations, measured in ground truth periods ($\sigma_{dir}$), was lower for the present method (0.05) than for {\bf mindtct} (0.39).

The ground truth periods were systematically smaller than the estimated periods, which can be explained by the fact that replacement noise increases the observed frequency. Quantitatively, our method exhibited a small bias of 0.09 in period estimation, with a remarkably low standard deviation of 0.03, indicating stable performance.

Detection instability primarily affects singularities with smaller periods under increased noise, although graceful degradation is less systematic for {\bf mindtct}, as illustrated by the false (pink) singularity near the Bottom-Right corner (Fig. \ref{fg:minutia_synt_mm_sp_noise_infull_agr}). This observation confirms that fingerprints of young children are more vulnerable to noise than those of adults at the currently practiced imaging resolution (510 dpi).

\section{Conclusions}

\begin{compactitem}
 \item The ST analytic signal $ f_A $ is continuous and isotropic. This representation is useful for image enhancement, as demonstrated in $\TWD$. Its extension to $ \ND $ is straightforward.
 
 \item The vector representation of the phase gradient—the instantaneous frequency vector,
is continuous and isotropic in magnitude. This property holds in $ \ND $ as well.

 \item The tensor representation of the phase gradient
is continuous and isotropic in both magnitude and orientation. It retains the same information as the instantaneous frequency vector. This property also holds in $ \ND $.

 \item The ST representation of the phase gradient enables singularity detection, and results confirm its practical relevance for wave physics, including fingerprint processing and optical fringe patterns.

 \item We show that images can be generated with full ground-truth
   control of singularities and high fidelity across diverse
   properties—type, direction, period, singularity number per scale, and a dense range of ridge scales with linearly varying wave periods—while ensuring no additional singularities, a key novelty of our modeling.
\end{compactitem}

{
    \small
    \bibliographystyle{ieeenat_fullname}
    \bibliography{bibl}
}

\clearpage
\renewcommand{\thetitle}{\strut\titletext}
\setcounter{page}{1}
\maketitlesupplementary

\renewcommand{\theequation}{S\arabic{equation}}
\setcounter{equation}{0}

\renewcommand{\thefigure}{S\arabic{figure}}
\setcounter{figure}{0}

\renewcommand{\thetable}{S\arabic{table}}
\setcounter{table}{0}

\renewcommand{\thesection}{S\arabic{section}}
\setcounter{section}{0}

\hypersetup{pageanchor=true}
\renewcommand{\thepage}{S\arabic{page}} 
\setcounter{page}{1}

\renewcommand{\theHsection}{S\arabic{section}}
\renewcommand{\theHequation}{S\arabic{equation}}
\renewcommand{\theHfigure}{S\arabic{figure}}
\renewcommand{\theHtable}{S\arabic{table}}

\emph{This document aims to ensure reproducibility. It provides a
summary of the models used to generate the test images, implementation
details of the methods presented in the paper, and additional
experimental results. The corresponding software will be made
available once the paper is published.}

\section{Test Images without Singularities}

Ground truth on local structure, or high-quality reference images, is difficult or impossible to obtain in many practical applications, for example in fingerprints collected at crime scenes. One goal of this section is to enable the automatic generation of large numbers of images that contain realistic local structure in terms of orientation, scale, and phase, with or without singularities, and with precise ground truth for evaluating algorithms, e.g. with respect to isotropy in orientation or scale, behaviour under contamination, or conditions where high-quality data are not readily available.

\begin{figure}[t]
\begin{center}
\fbox{%
\includegraphics[width=0.96\columnwidth]{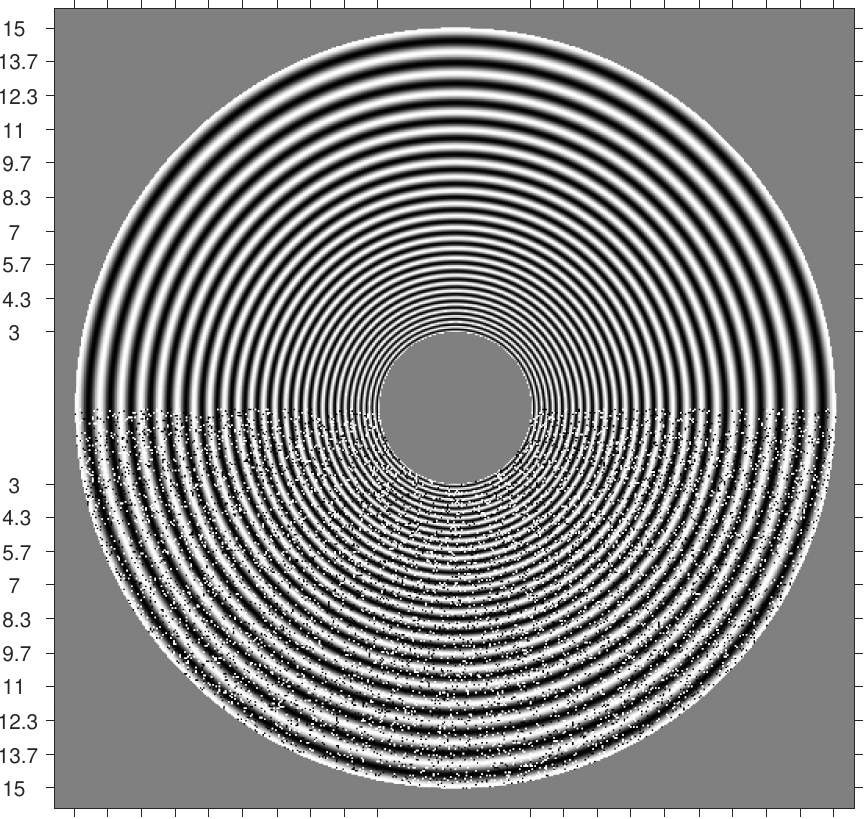}%
}
\end{center}
\caption{\label{fg:fmtest_logz_v4_0_419_2_094_legend} {\bf FM test
image} of size 512×512 pixels, with Salt \& Pepper replacement contamination in one half. Inside the circular region, the wave period $T$ (in pixels) varies linearly with the radius $r$, ranging from $3$ to $15$ pixels. This configuration simulates a spatial-frequency gradient from the center to the edge. The vertical legend indicates the instantaneous radial period, which can be directly read along the vertical midline of the image.}
\end{figure}

The phase factor $ C $ in $\cos(C\log|\br|)$ ((\ref{eq:cos_phi_L_our})) determines which frequencies can be rendered in the resulting FM test image, Fig.~\ref{fg:fmtest_logz_v4_0_419_2_094_legend}. Conversely, if the user sets $ r_{\max} $, representing the outer radius of the circle in the test image, and its scale, $ \omega(r_{\min}) $, then $ C $ can be determined, since the scale is the gradient norm, $ \omega = \frac{C}{r} $.
\begin{equation}
C = \omega_{\min}\cdot r_{\max}
\end{equation}

The user may then freely set the maximum absolute frequency $ \omega_{\max} $, provided it is less than $ \pi $, the Nyquist frequency. This implies that the inner radius is determined as $ r_{\min} = \frac{\omega_{\min}}{\omega_{\max}} r_{\max} $. Hence, the user’s setting of $ \omega_{\min} $, $ \omega_{\max} $, and $ r_{\max} $ implies that, as $ r $ varies in
$[r_{\min}, r_{\max}] = [\frac{\omega_{\min}}{\omega_{\max}} r_{\max}, r_{\max}]$, the wave period varies exactly as
\begin{equation}
T = \frac{2\pi}{C} r  = \frac{2\pi}{\omega_{\min} r_{\max}} r
\end{equation}
i.e. linearly with $ r$ over this interval.

\begin{figure*}[t]
    \centering
    \setlength{\tabcolsep}{2pt} 
    \renewcommand{\arraystretch}{0} 
    \begin{tabular}{ccc}
        \includegraphics[width=0.32\textwidth]{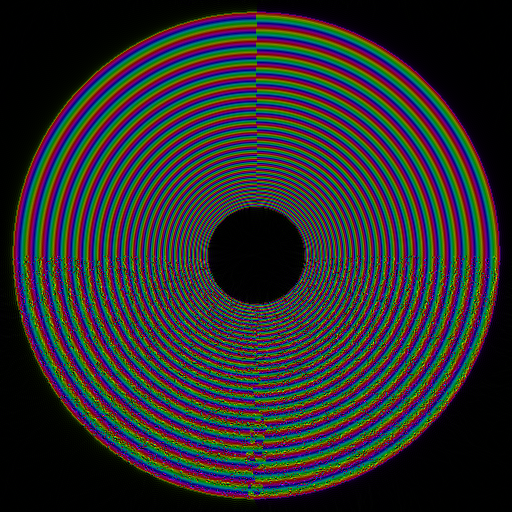} &
        \includegraphics[width=0.32\textwidth]{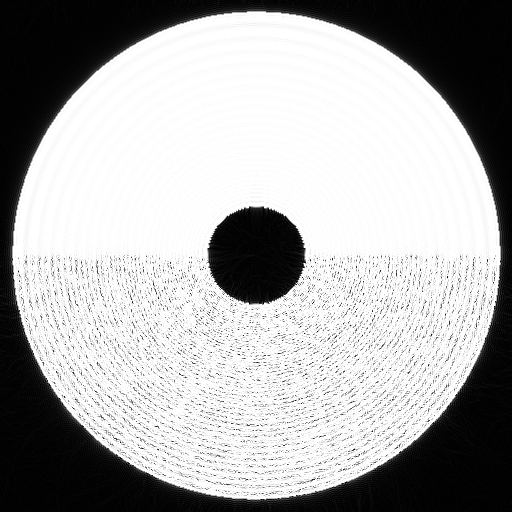} &
        \includegraphics[width=0.32\textwidth]{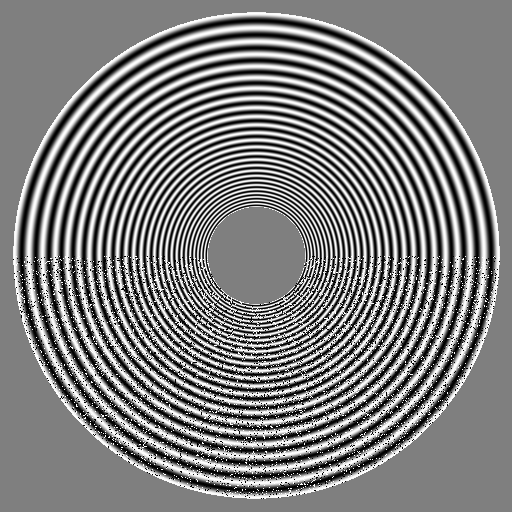} \\
        
        \includegraphics[width=0.32\textwidth]{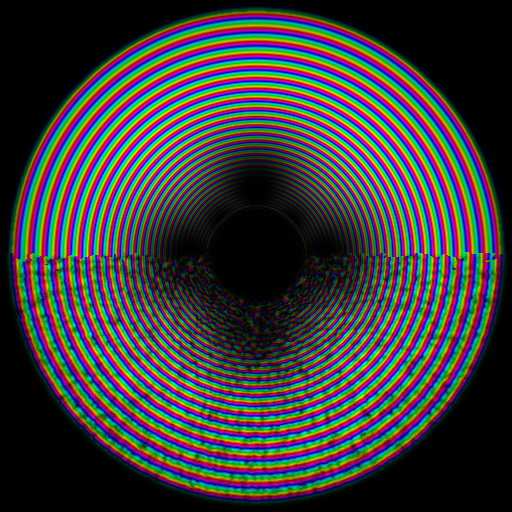} &
        \includegraphics[width=0.32\textwidth]{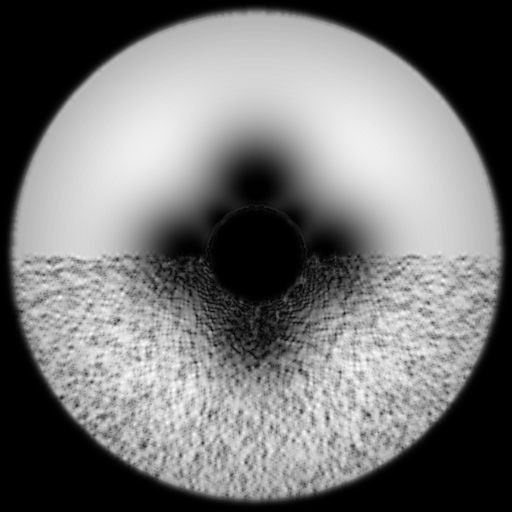} &
        \includegraphics[width=0.32\textwidth]{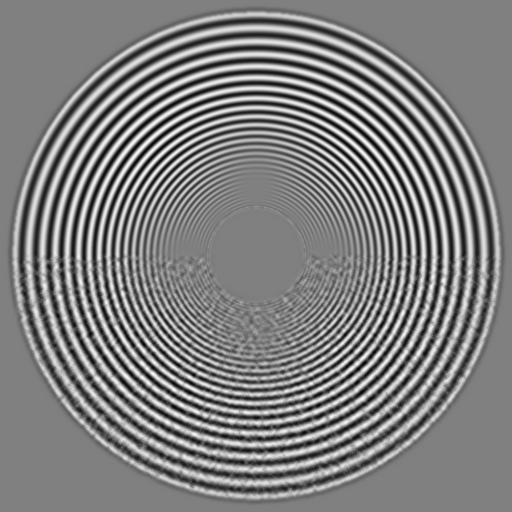} \\
        
        \includegraphics[width=0.32\textwidth]{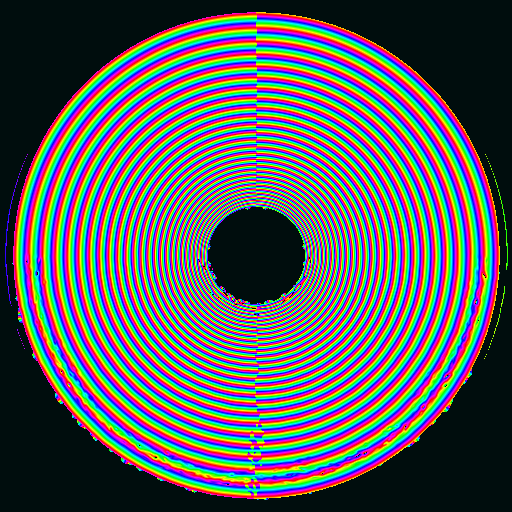} &
        \includegraphics[width=0.32\textwidth]{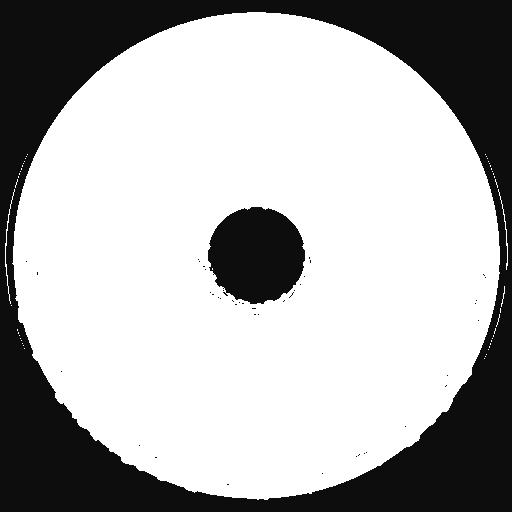} &
        \includegraphics[width=0.32\textwidth]{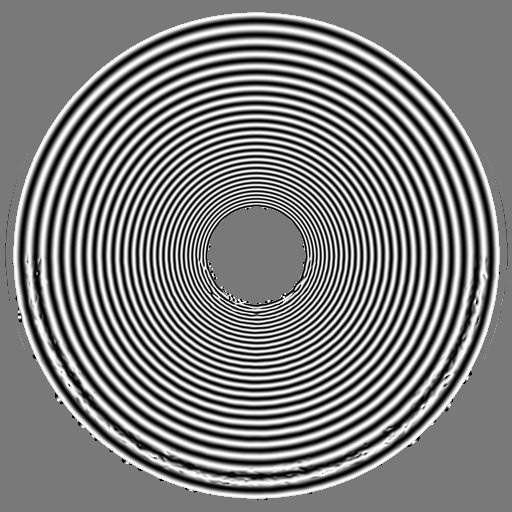} \\
    \end{tabular}
\caption{\label{fg:st_sa_real} {\bf Comparison} of three analytic-signal methods. Top row: Larkin–Felsberg monogenic signal. Middle row: Unser’s monogenic signal based on a spline–wavelet pyramid. Bottom row: analytic signal estimated using the proposed structure-tensor phase method. Columns, left to right: analytic signal (hue = phase, brightness = amplitude), amplitude of the analytic signal, and enhanced image (the real part $ \Re(AS) $).}
```

\end{figure*}

Fig.~\ref{fg:st_sa_real} shows a grid of images illustrating the
results of three methods for computing the analytic signal: top row
\cite{larkin2001natural}\cite{Felsberg01}, middle row\footnote{“The
color image in the middle row also contains separator-border
discontinuities, but these are less visually prominent because they
occur horizontally and are partly masked by the contamination
border. In contrast to the other two rows—where the noise border was
introduced deliberately to keep the border discontinuities visible—the
reduced prominence in the middle row result from the original
authors’ choice of analytic-signal half-space together with our
placement of the contamination region. Nonetheless, the discontinuities
remain discernible upon closer inspection even for the middle row.
} \cite{Unser09}, and our method (bottom row).

These support the following conclusions.
i) The real part of the monogenic signal cannot, by construction, provide a 1-D approximation of the original signal for enhancement; its real part is simply the original image.
ii) The monogenic signal applied to a spline–wavelet pyramid improves enhancement but exhibits a clear orientation bias, even at the pyramid level that best approximates the original image.
iii) Our structure-tensor-based phase estimation clearly improves upon the state of the art in analytic-signal estimation in the presence of nonlinear contamination (random replacement of pixel values).

The magnitude of the analytic signal should be 1 (i.e. constant) when the local input is a sinusoid. This holds for the first and third rows within the circular area in the top half. However, in the bottom half, only our method (the third row, middle image) preserves this important property under contamination.

Regarding the amplitude in the middle row, the contamination affects it less than in the first row, but it is still clearly more affected than in the last row and is orientation biased in comparison.

That our method objectively resists contamination better can be
explained by the fact that the underlying analytic signal is a 1-D
approximation of the local input; it does not assume this 1-D  structure, unlike state-of-the-art methods relying on the monogenic signal.

\section{Phase Singularities and Ground Truth}

In addition to traditional ground truths—location, direction, and singularity
type—we require control over absolute frequency (scale). We also require control
over the amount and nature of contamination/noise that degrades image quality,
so that we may simulate the effects of adverse imaging conditions. Finally, the
generated images should allow for efficient visualization and human inspection.

While our work is inspired by the phase-based singularity models of
\cite{sherlock93,larkin07synth}, these also emphasize the importance of
orientation maps, which form a significant part of the analytic phase and can be
efficiently obtained via the structure tensor (ST). We extend these models from
multiple singularities to \emph{multiple steerable singularities}, enabling
explicit synthesis of their properties in addition to localization.

Fig.~\ref{fg:minutia_synt_mm_sp_noise_infull} shows one of many possible
synthesized images with known ground truth at every pixel. In the examples used
here (image size $512 \times 512$), a total of 70 singularities—both
\emph{bifurcation} and \emph{ridge-end} types—are embedded in a background wave
whose instantaneous period (i.e., scale) varies continuously across a wide
range. Singularity locations and directions span the full image domain and the
full $2\pi$ range. Random salt-and-pepper replacement (SPR) noise is applied to
simulate varying degrees of degradation. Many parameters can be varied
continuously without losing control, and importantly no additional unintended
singularities are introduced.

We begin with a single-singularity model (Eq.~\ref{eq:minutia_model}), and then
extend it to a constellation of singularities.

The model expresses the phase as a sum of a linear component (planar wave) and a
nonlinear component (phase singularity):
\begin{equation}
\Phi(\br)
= \Phi_L(\br) + \Phi_P(\br)
= \bom^\T\br + \arg\!\left(\frac{\tilde\br}{i\tilde\bom}\right),
\label{eq:minutia_model}
\end{equation}
where $\bom,\br \in \mathbb{R}^2$. The operator $\tilde{\cdot}$ converts a
$2$-D vector into its complex representation; e.g.\ $\br = (1,1)^\T$ yields
$\tilde\br = 1+i$.

The first term $\Phi_L(\br)$ (linear phase) generates a local planar wave whose
frequency vector is $\bom$. Later, $\bom$ may vary slowly with $\br$, so that it
is approximately constant in any small neighborhood.

The second term $\Phi_P(\br)$ (polar phase) produces a continuous phase increase
(modulo $2\pi$) around the origin, here positioned at the evaluation point to
avoid symbol clutter. Dividing $\tilde{\br}$ by $i\tilde\bom$ initializes the
phase increase from the iso-curve direction of the local ridge. Combining the
two terms yields a singularity of ridge-end type with direction
$\arg(i\tilde\bom)$, in accordance with ISO conventions for fingerprint
minutiae, with $|\tilde\bom|$ giving the absolute frequency.

Other singularity directions can be obtained by rotating the local coordinate
frame, and other scales by adjusting $\|\bom\|$. A ridge-end singularity is
similar to Eq.~\eqref{eq:minutia_model} but phase-translated as
$\Phi' = \Phi - \pi$.

A synthetic image $I$ containing a singularity at the origin is produced by
taking the real part of the analytic signal:
\begin{equation}
I(\br) = \Re{\exp\big(i\Phi(\br)\big)},
\label{eq:reexpfi}
\end{equation}
where the iso-gray curves of $I$ coincide with the level sets of $\Phi$.

For a set of desired singularities at locations $\br_j$ with types
$t_j \in \{0,1\}$ (0 = bifurcation, 1 = ridge-end), a synthetic fingerprint-like
image is generated by the additive polar-phase model:
\begin{equation}
\Phi(\br)
= \Phi_L(\br)
+ \sum_j (-1)^{t_j}
    \arg\!\left(\frac{\tilde\br - \tilde\br_j}{i\tilde\bom_j}\right),
\label{eq:combined_minutia_model}
\end{equation}
where $\nabla\Phi_L(\br)$ varies slowly and may be treated as constant in the
neighborhood of $\br_j$, i.e.\ $\nabla\Phi_L(\br) \approx \bom_j^L$. When
aggregated, singularities must be described relative to a common origin rather
than their individual ones, explaining the translations by $\tilde\br_j$.

The singularity direction must remain orthogonal to the local wave vector
$\bom_j = \nabla\Phi(\br_j)$, assumed constant in a neighborhood of $\br_j$.
This is ensured by the division by $i\tilde\bom_j$. Note that $\bom_j$ is
constant but generally differs from $\bom_j^L$ (defined above).

The instantaneous wave vector is then
\begin{eqnarray}
\bom(\br)
 &=& \nabla\Phi(\br)
  = \nabla\Phi_L
    + \sum_k (-1)^{t_k}
      \nabla\arg\!\left(\frac{\tilde\br - \tilde\br_k}{i\tilde\bom_k}\right)
      \nonumber \\
 &=& \bom^L
    + \sum_k (-1)^{t_k}
      i\,\frac{\tilde\br - \tilde\br_k}{|\tilde\br - \tilde\br_k|^2},
\label{eq:grad_phase}
\end{eqnarray}
where the last line uses the fact that the gradient fields of polar phases are
identical for different singularity directions when expressed in local
coordinates. Sampling at $\br_j$ yields $\bom_j$ upon substitution of $\br =
\br_j$.

For the image in Fig.~\ref{fg:minutia_synt_mm_sp_noise_infull}, we used a
the logarithmic  function:
\begin{equation}
\Phi_L(\br)=C\,\log\|\br\|,
\label{eq:phi_L_our}
\end{equation}
 generating a slowly varying
linear phase in a sinusoid, e.g.  $ \cos(\Phi_L(\br) )$. 
Here $C$ controls the range of scales and the coordinate origin is the image
center. The corresponding $\bom_j$ follows from Eq.~\eqref{eq:grad_phase}:
\begin{equation}
\bom_j
= \bom(\br_j)
= C\frac{\br_j}{\|\br_j\|^2}
  + \sum_k (-1)^{t_k}
     i\,\frac{\tilde\br_j - \tilde\br_k}{|\tilde\br_j - \tilde\br_k|^2},
\label{eq:bom_j_our}
\end{equation}
using $\nabla\Phi_L = C\br_j/\|\br_j\|^2$. Once $\Phi_L$ and $\bom_j$ are known,
the full phase and image are synthesized\footnote{Software will be made
publicly available after publication.} by substituting
(\ref{eq:bom_j_our}) and (\ref{eq:phi_L_our}) into
(\ref{eq:combined_minutia_model}) and then applying~(\ref{eq:reexpfi}). Noise
contamination is applied afterward.

\begin{figure}[t]
\begin{minipage}[t][][t]{\columnwidth}
\fbox{
\includegraphics[width=0.96\columnwidth]{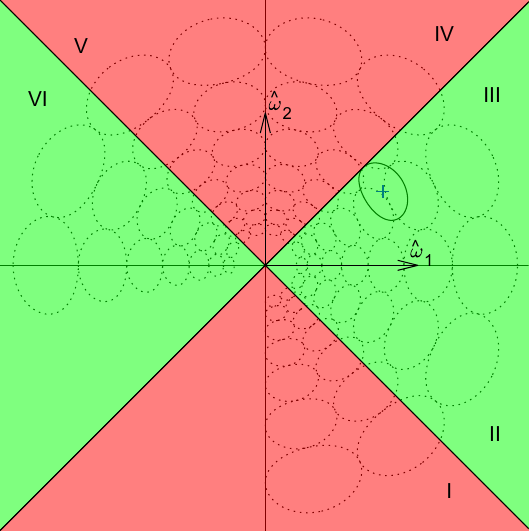}
}
\caption{\label{fg:gd_logz2_4_}  FT in $ \TWD$ with two different
   analytic
  signal representations of the same signal, 1-IV being the first with $ \bn=\hat\bom_1
  $ and III-VI  being the second with $ \bn=\hat\bom_2$.  Green II-III marks
  the region 
  that is safest with $\bn=\hat\bom_1$. 
  Iso curves of Gabor filters  
for  inputs with different $ \bom$ are in black.
}
\end{minipage}
\begin{minipage}[t][][t]{\columnwidth}
\fbox{
\includegraphics[width=0.96\columnwidth]{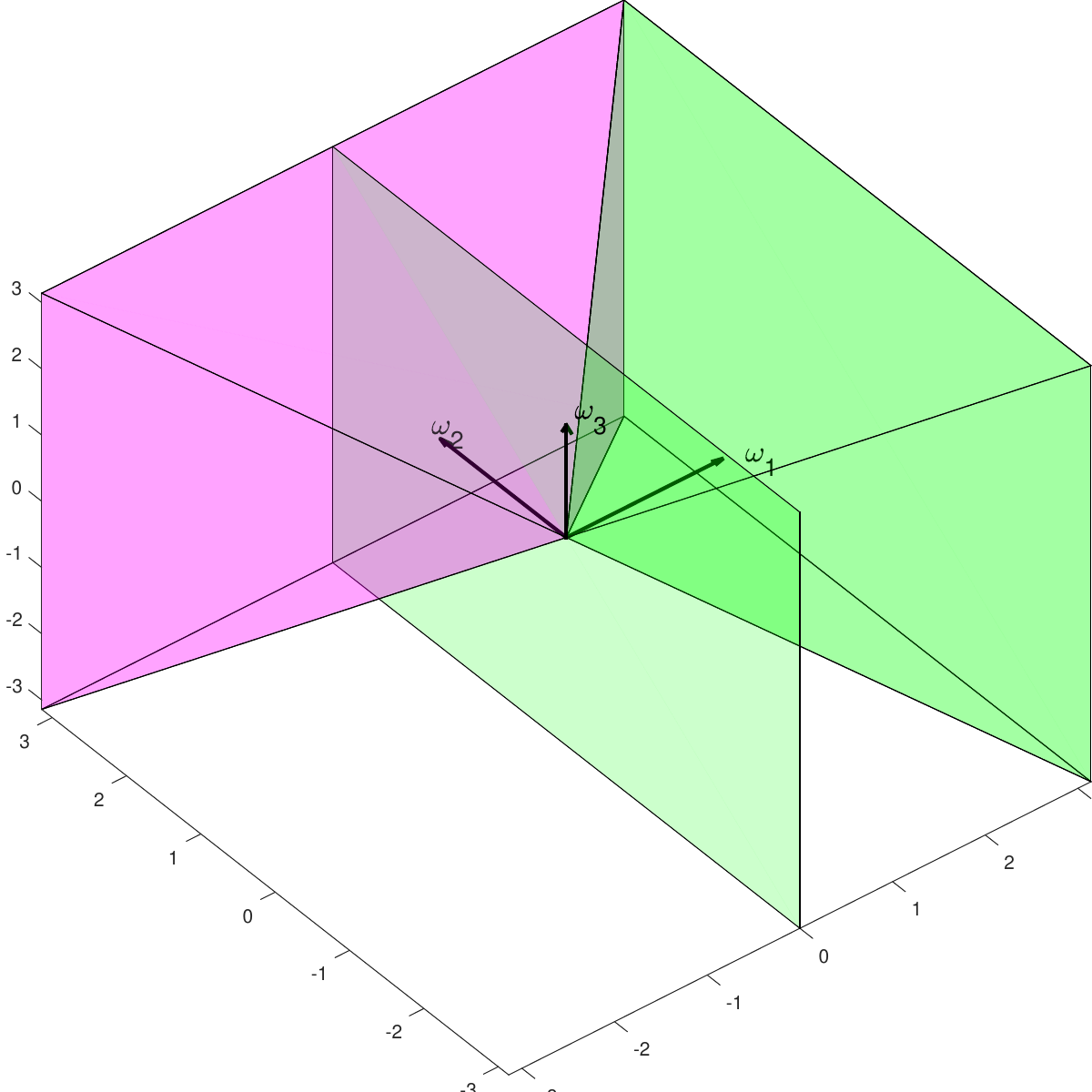}
}
\caption{\label{fg:pyramids_in_cube}  FT in $ \THD$ with its 
  reliable pyramid  (green) under the assumption
  $\bn=\hat\bom_1 $ defining the (green) hyperplane. Dito pyramid (pink), but for the
  alternative $\bn=\hat\bom_2 $ (hyperplane not drawn).}
\end{minipage}
\end{figure}

\section{Gradient of phase in N-D \label{sc:gradph}} 
\subsection{Direct Gradient} 
\begin{figure}[t]
\begin{minipage}[t][][t]{\columnwidth}
  \fbox{%
  \includegraphics[width=0.96\columnwidth]{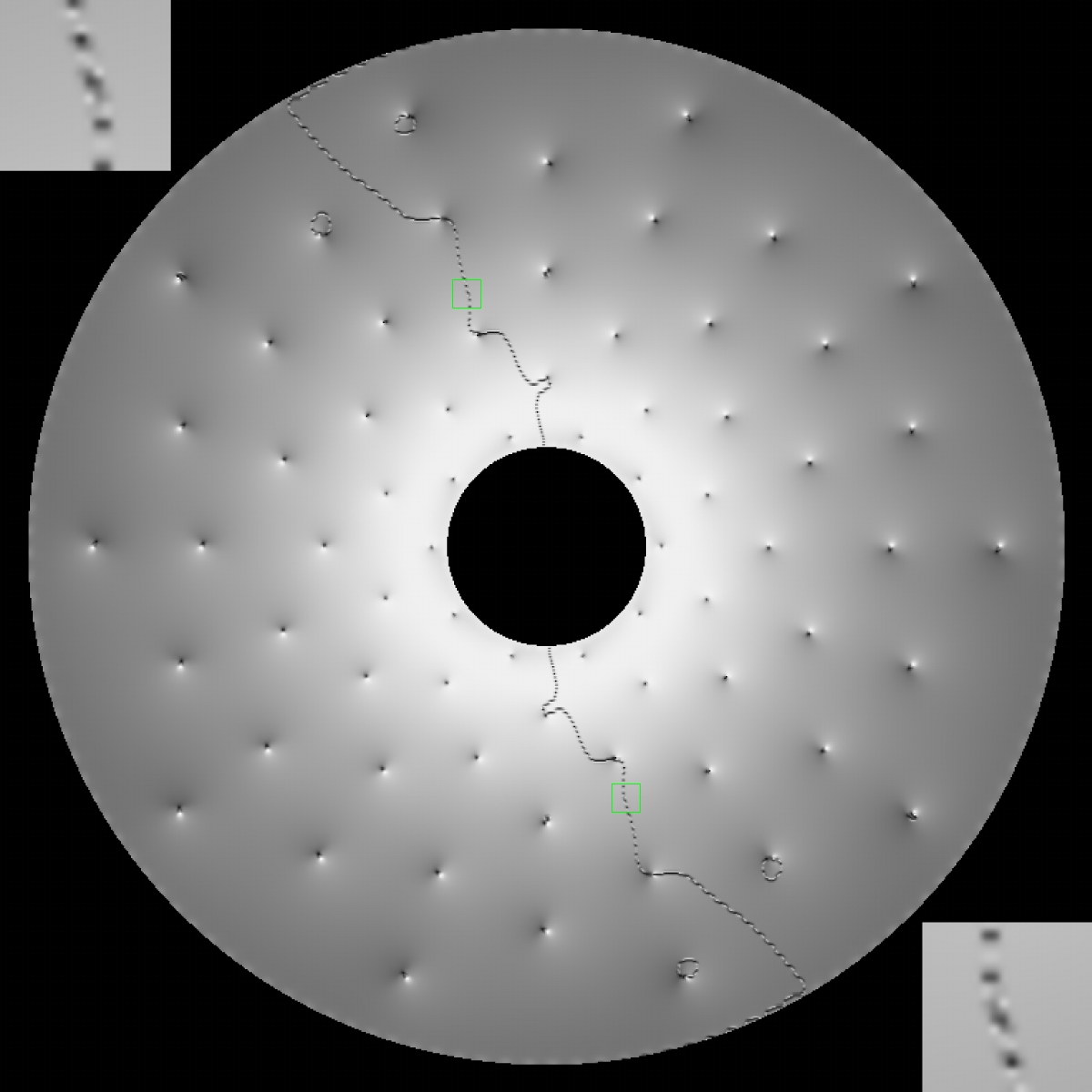}%
  }
  \caption{\label{fg:ph_mag_gradient}
  Magnitude of the ST phase gradient, $\|\nabla\Phi^{\bST}\|$, direct
  implementation. Zoomed squares are inset at the nearest corners.}
\end{minipage}
\end{figure}

 Real and
  imaginary parts of Gabor filter responses,
\begin{equation}
z_{\bom(f)}
=u+iv=\exp(i\Phi^{\bST}) \label{eq:stdp_normalized}
\end{equation} 
 are differentiable, where  the amplitude  $ |z_{\bom(f)}|$  is
 set to 1, without loss of generality. The dependence of $ f,u,v, \Phi^{\bST} $, and  $ z_{\bom(f)}$  on
$\br $ is not spelled out to reduce clutter. Consequently, the
 Gradient is given by
\begin{equation}
\nabla\Phi^{\bST}= -v\nabla u +u\nabla v \label{eq:gradphi} 
\end{equation}

 Fig.  \ref{fg:ph_mag_gradient} shows the magnitude of  gradients
 $\|\nabla\Phi^{\bST} \| $ according to (\ref{eq:gradphi}), evidencing
 continuity at points of  circular
 artifacts. This is because the
phase gap at such an  $ \br$ 
 is  $2\pi $ which is in turn  equivalent to multiplying $z_{\bom(f)}
 $  with $\exp(i2\pi)=1 $, leaving 
$ u, v $,  in $ \nabla u, $  and $\nabla v $  and thereby, in $ \| \nabla v  \|$   unaltered.

In mid vertical portion, (some zoomed-in), local images are planar waves obeying  $\bom^\T \bn=0 $. As evidenced in the image, hyperplane artifacts  survive
 standard implementation of derivation operator\footnote{convolution with  gradient
 of a Gaussian}   defying the
 intuition from $\OND $.  The
 phase difference now is no longer an integer multiple of $2\pi $ but
 $2|\bom^\T\br| $, which is noticeable in real, and imaginary  parts of
 $ z_{\bom(f)}(\br) $ 
 and
 their gradients, as a  zig-zag patterned artifact wave.  Next, we
 suggest remedies.

\subsection{Continuous magnitude of compound
  gradients   } 
We detail compound implementation in  $\TWD$,  Fig. \ref{fg:gd_logz2_4_},  for pedagogical reasons.  The idea  will be
extended to $\ND $ subsequently.  Let $ \hat \bom_1$,
$ \hat \bom_2$, be basis vectors of  2D FT domain.  We assume that the walking direction vector
is, $ \bn=\hat 
\bom_1$. Then the specific  input  sinusoid,  $ f(\br)=\cos (
\omega\hat\bom_1^\T\br)$, with $ \bom$ located in $\pm \pi/4 $ around $\bom_1
$ will produce the most reliable phase. 
By contrast, if the walking direction   $ \bn=\hat\bom_2$ would have been
assumed, the  $\bom $ of the input sinusoid  in region $ \pm \pi/4 $
around $ \bom_2$ 
would produce the most reliable phase.

By using two different representations, 
 $\bn\!=\!\hat\bom_1 $, and  $\bn\!=\!\hat\bom_2 $ respectively,
one can 
obtain two phases  $\Phi_1^{S}  $ and $\Phi_2^{S}  $ having
different  reliabilities. Each
signal representation would then have 
its own half-space of FT where the respective  analytic function $
z_{\bom(f)} $  is non-zero (and zero). This requires the additional knowledge of  $\Phi^{\bST}_2 $,
which can be construed from $\Phi^{\bST}_1$, 
\begin{equation}
\Phi^{\bST}_2(\br)   = \sgn(\hat\bom_2^\T\bom(\br)) \Phi^{\bST}_1(\br)
\label{eq:grad_ph_comp_1} 
\end{equation} 
One can then expect that  $  \| \nabla \Phi_1^{S} \|= \|
\nabla \Phi_2^{S} \|=\omega  $ will be continuous when $\bom $ is at
a comfortable distance from the applicable hyperplane.

After defining the index $ J$ 
\begin{equation}
 J(\br)=\argmax_j \{ |\hat\bom_j^\T\bom(\br)|\}_{j=1}^{N=2} 
\end{equation} 
identifying the most  reliable phase,
we can apply  gradient
to $\Phi^{\bST}_1 $, and $\Phi^{\bST}_2 $,  yielding  the
  compound gradient 
\begin{equation}
\nabla\Phi^{\bST} (\br) = \sgn(\hat\bom_J^\T\bom(\br))\nabla\Phi^{\bST}_J(\br) 
\label{eq:grad_ph_comp_2} 
\end{equation} 
where $\sgn$ is the sign function. The
direction needs sign-rectification  to assure that  $\bom $ (computed
by ST) points  into the non-zero zone when the corresponding $ \nabla \Phi^{\bST}_2 $
will be from the reliable zone of  $  \Phi^{\bST}_2 $.

Indeed, as can be seen in Fig. \ref{fg:phsr_Daleth_COMP_omest} the intensity of the color image is continuous, and the intensity represents the magnitude of the gradient (to be precise $ \|\nabla \Phi^{\bST}\|^2$) which was having discontinuity in the direct implementation of gradients, Fig. \ref{fg:ph_mag_gradient}. In other words, the zig-zag discontinuity in neighborhoods having hyperplane discontinuity has been eliminated by compound gradient implementation. The color in the image will be precised in Sec \ref{sc:cls}. 

Compound gradient implementation and conclusions above can be generalized to $\ND $ in a
straight forward manner. Assuming an analytic signal
having $\bn=\hat\bom_1 $,  the region
of most reliable phase in the FT space is a
hyper-pyramid around $\bn $, green "Egyptian'' pyramid in  Fig. 
\ref{fg:pyramids_in_cube}.    The least reliable region is  the remainder of the
non-zero zone, i.e.   $ N-1$  half hyper-pyramids  
between  the reliable hyper-pyramid and  the hyperplane  $
\bn^\T\bom=0$, marked green in the figure. 
Phase being $\Phi^{\bST}_1  $  under the presumption (of $
\bn=\hat\bom_1 $),  $N-1 $   alternative phases $\Phi^{\bST}_2, \cdots \Phi^{\bST}_N  $  can be
construed from it   based on
 respective  presumptions on walking directions,  $\bn=\hat\bom_2,
\cdots \bn=\hat\bom_3 $  (without   filtering):
\begin{equation}
\Phi^{\bST}_j(\br)  =   \sgn(\hat\bom_j^\T\bom(\br)) \Phi^{\bST}_1(\br) \quad
j\in 2\cdots N
\end{equation} 
 each having its own reliable
hyper-pyramid. Such a  pyramid is marked as pink in $\THD $ in the figure
for $ \Phi^{\bST}_2 $.

A compound phase gradient is then sewn together by cherry picking among phase   gradients of
reliable pyramids. A practical way of  achieving this is by  using  the index $ J$  
\begin{equation}
J(\br)  =
\argmax_j(\{|\hat\bom_j^\T\bom(\br)|\}_{j} )\qquad j,J\in
1\cdots N \label{eq:pyr_index} 
\end{equation} 
It identifies  the most reliable hyper-pyramid or its mirror in the 
 zero-zone, in which  $\bom(\br) $
is. The ambiguity when max is
not unique (at hyper-pyramid boundaries), 
is resolved  e.g. by seting J as min of (two)  $ j$s in the cause. Note that $\bom(\br) $ is always located in the non-zero zone
of the 
FT partition produced by  the $\bn=\hat\bom_1 $ assumption, albeit the location can
be in its non-reliable zone. In the latter case, the
location   can be in
the zero zone of another partitioning of the FT space. 

 A phase gradient
selection, the  analog of  (\ref{eq:grad_ph_comp_2}), 
decided by  $ J$  produces finally the compound gradient. 
\begin{equation}
\nabla\Phi^{\bST} (\br) \!  = \!\sgn(\hat\bom_{J(\br)} ^\T\bom(\br))
\nabla\Phi^{\bST}_{J(\br)} (\br)  \label{eq:grad_ph_comp_3} 
\end{equation} 

Apart from  book-keeping cost of constructing $\sgn(\bom_j^\T
\bom(\br)) $,  and  $ J(\br)$, 
compound gradients can be obtained  at  the same 
 cost as  direct implementation. This is because  $J(\br) $  can be  used to
 compute not more than one  $\nabla\Phi^{\bST}_{J(\br)} (\br)  $  per
 point $ \br$. Book-keeping cost is at  the order of $N^2 $ arithmetic
 operations per point, if  projections  $\bom_j^\T\bom $ are computed
 directly. However, rather than projections,
 look-up tables on components of $\bom $ can be used to
 decide signs and orderings of $\bom_j^\T\bom $, 
 marginalizing the overhead.

 In summary, the  compound phase gradient 
\begin{equation}
 \{\bS^D , \bS^S, \Re{\exp(i\Phi^{\bST})}, \|\nabla\Phi^{\bST} \|\}   
\end{equation}
 is  a continuous and isotropic representation in {\em magnitude}. However, the direction of phase gradient remains discontinous accross neighborhoods with wave vectors on the Analytic Signal hyperplane\footnote{Not illustrated  since it follows from $\nabla\Phi^{\bST}=\pm \bom $  at hyperplane discontinuity. }. This means that while $\|\nabla\Phi^{\bST} \|$ can be smoothed when implemented via compound gradients,  $\nabla\Phi^{\bST} $ cannot.

\subsection{\label{sc:cls} Structure Tensor (ST) of  Phase}
 Compound gradient achieves  continuity of  $\|\nabla\Phi^{\bST}\|  $  but falls short of providing
it for the   direction. This is
remedied  by   CST of phase being the (infinitesimal) Structure Tensor
 applied to ST Phase, 
\begin{equation}
\nabla\Phi^{\bST}\nabla^\T\!\Phi^{\bST} 
\end{equation} 
  In 2D, this  is equivalent  to complex squaring
 of the (compound) gradient \cite{bigun87london}, when it is seen as a
 complex number  i.e. $(\tilde\nabla\Phi^{\bST})^2 $. The  tensor is illustrated  in
  Fig. \ref{fg:phsr_Daleth_COMP_omest},  with hue
 depicting the direction of ``squared gradients''. 
That neither magnitude nor direction  suffer from hyperplane
 artifacts  is
 evidenced by the image. Additionally, this  tensor can be
 averaged (as opposed to  $ \nabla \Phi^{\bST}  $ vectors),  suggesting better estimates of
 the local direction {\em and } the scale, $\| \bom\| $ \cite{bigun16lasvegas},\cite{lindeberg98}.

\end{document}